%% file: main.tex
\documentclass[journal]{IEEEtran}
\usepackage{amsmath,amsfonts, amssymb, amsthm}
\usepackage{algorithmic}
\usepackage{algorithm}
\usepackage{array}
\usepackage{textcomp}
\usepackage{stfloats}
\usepackage{url}
\usepackage{verbatim}
\usepackage{graphicx}
\usepackage{cite}
\usepackage{multicol}
\usepackage{multirow}
\usepackage{colortbl}

\usepackage{booktabs,multirow,bm}
\usepackage{tabularx}

\usepackage{xcolor}
\usepackage{lipsum}

\ifCLASSOPTIONcompsoc
  \usepackage[caption=false,font=normalsize,labelfont=sf,textfont=sf]{subfig}
\else
  \usepackage[caption=false,font=footnotesize]{subfig}
\fi

\input{math_commands}

\begin{document}
	%
	\title{Toward Real-Time Edge AI: Model-Agnostic Task-Oriented Communication with Visual Feature Alignment}
	%
	%
	%
	\author{Songjie~Xie,~\IEEEmembership{Graduate~Student~Member,~IEEE}, Hengtao~He,~\IEEEmembership{Member,~IEEE}, Shenghui~Song,~\IEEEmembership{Senior~Member,~IEEE}, Jun~Zhang,~\IEEEmembership{Fellow,~IEEE}, and Khaled~B.~Letaief,~\IEEEmembership{Fellow,~IEEE}
		\thanks{The authors are with the Department of Electronic and Computer Engineering, The Hong Kong University of Science and Technology, Hong Kong (e-mail: sxieat@connect.ust.hk, \{eehthe, eeshsong, eejzhang, eekhaled\}@ust.hk). (The corresponding author is Hengtao He.)}
	}

	\maketitle
	
\begin{abstract}
Task-oriented communication presents a promising approach to improve the communication efficiency of edge inference systems by optimizing learning-based modules to extract and transmit relevant task information. However, real-time applications face practical challenges, such as incomplete coverage and potential malfunctions of edge servers. This situation necessitates cross-model communication between different inference systems, enabling edge devices from one service provider to collaborate effectively with edge servers from another. Independent optimization of diverse edge systems often leads to incoherent feature spaces, which hinders the cross-model inference for existing task-oriented communication. To facilitate and achieve effective cross-model task-oriented communication, this study introduces a novel framework that utilizes shared anchor data across diverse systems. This approach addresses the challenge of feature alignment in both server-based and on-device scenarios. In particular, by leveraging the linear invariance of visual features, we propose efficient server-based feature alignment techniques to estimate linear transformations using encoded anchor data features. For on-device alignment, we exploit the angle-preserving nature of visual features and propose to encode relative representations with anchor data to streamline cross-model communication without additional alignment procedures during the inference. The experimental results on computer vision benchmarks demonstrate the superior performance of the proposed feature alignment approaches in cross-model task-oriented communications. The runtime and computation overhead analysis further confirm the effectiveness of the proposed feature alignment approaches in real-time applications.
\end{abstract}
	

\input{secs/1_intro}
\input{secs/3_system}

\input{secs/4_method_1}

\input{secs/4_method_2}

\input{secs/5_exp}

\input{secs/6_ending}

\appendices
\input{secs/7_appendix}

\bibliographystyle{IEEEtran}
\bibliography{IEEEabrv,ref}

\end{document}

%% file: math_commands.tex
\newcommand{\x}{\mathbf{x}}
\newcommand{\z}{\mathbf{z}}
\newcommand{\y}{\mathbf{y}}

\newcommand{\hz}{\hat{\mathbf{z}}}
\newcommand{\tz}{\Tilde{{\mathbf{z}}}}
\newcommand{\hy}{\hat{\mathbf{y}}}
\newcommand{\bZ}{\mathbf{Z}}

\newcommand{\bE}{\mathbb{E}}

\newcommand{\KL}{\text{KL}}
\newcommand{\M}{\mathbf{M}}
\newcommand{\X}{\mathbf{X}}

\newcommand{\hMLS}{\hat{\mathbf{M}}_{\text{LS}}}
\newcommand{\hMSE}{\hat{\mathbf{M}}_{\text{MMSE}}}
\newcommand{\hMML}{\hat{\mathbf{M}}_{\text{MLP}}}

\newcommand{\bphi}{\boldsymbol{\phi}}
\newcommand{\btheta}{\boldsymbol{\theta}}

\newcommand{\bepsilon}{\boldsymbol{\epsilon}}
\newcommand{\ms}{\textrm{ms}}

\newcommand{\Sc}{\text{S}_{\text{C}}}

\DeclareMathOperator*{\argmax}{argmax}
\DeclareMathOperator*{\argmin}{argmin}
\DeclareMathOperator{\Tr}{Tr}

\newtheorem{definition}{Definition} 

\newtheorem{proposition}{Proposition}
\newtheorem{assumption}{Assumption}

\makeatletter

\newcommand{\Rmnum}[1]{\expandafter\@slowromancap\romannumeral #1@}
\makeatother

%% file: secs/1_intro.tex
\section{Introduction}
As artificial intelligence (AI) advances rapidly across diverse domains, the next generation of communication (6G) is poised to enable ubiquitous AI services~\cite{letaief2019roadmap, saad2019vision, du2023attention, 10422886}.
For the wide array of AI services ranging from Virtual/Augmented/Extended Reality (VR/AR/XR) to autonomous driving, real-time computer vision (CV) emerges as a pivotal technology to interpret visual information from images and videos for perceiving and modeling dynamic physical environments~\cite{voulodimos2018deep}.
However, the real-time performance of CV applications essentially depends on the tremendous visual data exchange, which brings an unprecedented burden on existing communication systems. To solve the problem, \emph{edge AI}~\cite{letaief2021edge} has been proposed, where AI applications are implemented in the network edge to eliminate the excessive latency incurred by routing data to the central cloud. It deploys AI models such as deep neural networks (DNNs) with the cooperation between edge devices and the nearby edge server to provide immersive services.

{While performing inference at the network edge offers numerous advantages, directly transmitting large volumes of visual data to the edge server can lead to significant communication overhead. To achieve low-latency edge inference, task-oriented communication has been proposed~\cite{gunduz2022beyond, shao2021learning, shi2023task, 10370739}, which focuses on extracting task-relevant information and transmitting it to the edge server for downstream tasks.}
As only task-relevant information is transmitted on the network edge, the communication overhead is significantly reduced.

Although edge inference with task-oriented communication provides a promising solution for real-time CV applications, the practical deployment brings additional requirements for system design.
One of the critical challenges is cross-model inference across various service providers, where an edge server from one provider performs inference using features transmitted from edge devices of other service providers. This will be further elaborated in the next section.

\subsection{Related Works and Motivations}
\subsubsection{Edge AI Inference}
Deploying AI from the centralized cloud to network edges has emerged as a promising solution to support ubiquitous AI services on wireless systems, attracting broad attention in industry and academia~\cite{mao2024green, xu2023edge, letaief2021edge}. 
Edge device-server co-inference endeavors to facilitate the seamless integration of ubiquitous AI applications within wireless networks by consolidating data access, processing, transmission, and downstream inference at the network edge~\cite{letaief2021edge}. In this framework, large DNN models are split between edge devices and edge servers to reduce the computation overhead by the device-edge collaboration~\cite{yang2020energy}. In partitioning the computation burdens between the edge devices and edge servers, features instead of raw data are transmitted, which leads to a trade-off between the communication and computation~\cite{shao2020communication}. To optimize this trade-off, adaptive model partitioning strategies were investigated via edge-device synergy, splitting learning, and dynamic neural networks~\cite{li2019edge,wu2023split}. 
Furthermore, as the communication overhead is determined by the dimensionality of feature vectors,  the incorporation of feature compression techniques can achieve low-latency edge inference~\cite{shao2020bottlenet++, shao2021learning}. These techniques effectively eliminate redundant dimensions, retaining only those dimensions essential for the task at hand. 
Thus, task-oriented communication has been proposed to shift system design from exact data recovery to task completion.

\subsubsection{Task-Oriented Communication} 
Recent advancements in task-oriented communications have embraced the integration of DNNs for the efficient extraction of task-relevant information and the successful execution of inference tasks. One key technique is the learning-based joint source-channel coding (JSCC)~\cite{bourtsoulatze2019deep, kurka2020deepjscc, farsad2018deep}, where DNNs in edge devices and servers are optimized in an end-to-end manner. The initial task-oriented communication scheme was developed based on the Information Bottleneck (IB) principle~\cite{tishby2000information}, aiming to transmit minimal information for the inference task~\cite{shao2021learning}. The IB-based task-oriented communication has also been extended to enhance the resilience against channel noise~\cite{xie2023robust} and shifts in the source distribution~\cite{li2024tackling}. Beyond single-device scenarios, task-oriented communication has been generalized to multi-device scenarios by adopting the principles of distributed IB~\cite{shao2022task, aguerri2019distributed} and maximal coding rate reduction~\cite{cai2024multi, cai2024end}. Besides, the principle of communication beyond the bit level is also adhered to and carried forward by recent studies of semantic communication focusing on the semantic aspect of information transmission~\cite{xie2021deep, xie2022task, zhang2022deep, dai2022nonlinear}. While task-oriented communication offers a promising avenue for low-latency edge AI systems, real-time deployment raises practical challenges, particularly regarding model compatibility across diverse systems.
\begin{figure}[t]
		\centering
		\includegraphics[width=8.5cm]{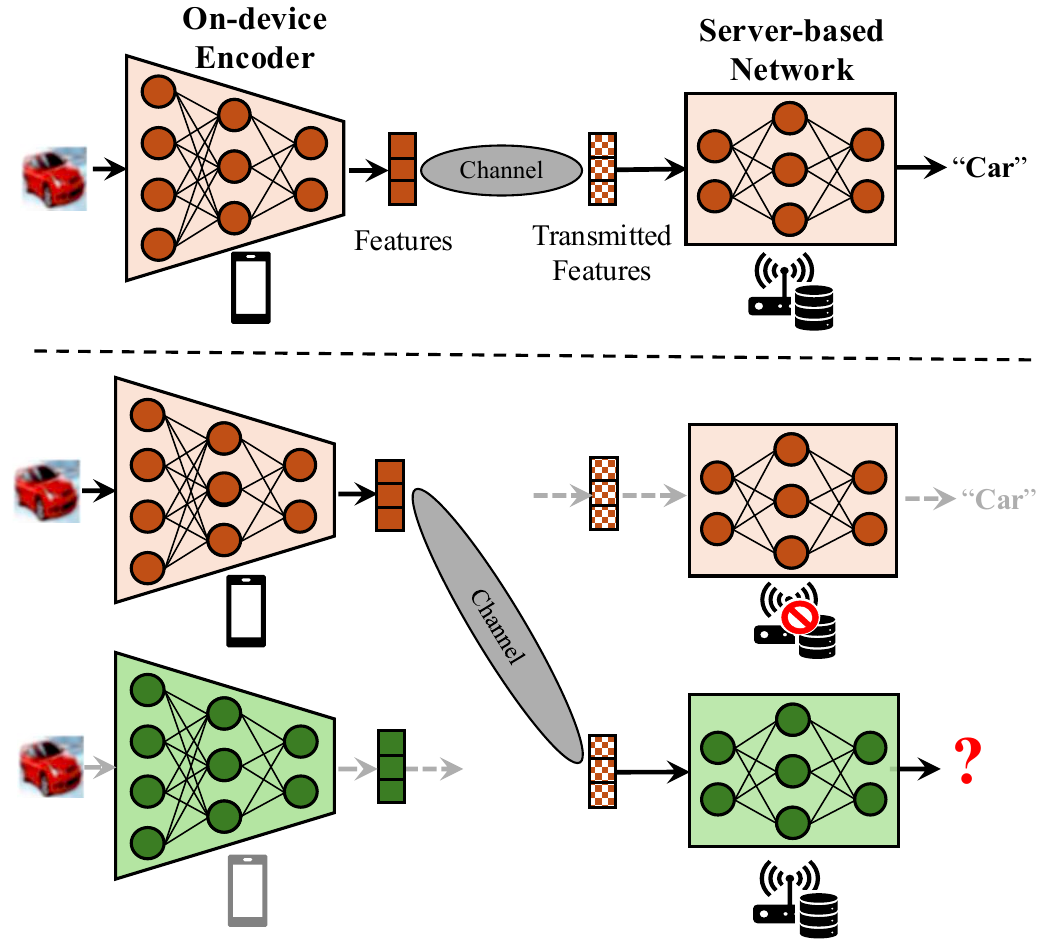}
		\caption{The illustration of the compatibility issue in cross-model task-oriented communications. 
		}
  \label{fig:intro}
\end{figure}

Although task-oriented communication has presented superior performance in edge inference, there are critical challenges when implementing task-oriented communications in practical edge AI systems. In particular, the proliferation of AI services necessitates collaborations with diverse service providers, where unique DNN architectures are trained independently due to privacy constraints and intellectual property considerations. The challenges in practical deployments, such as incomplete coverage and potential edge server malfunctions~\cite{wang2020convergence}, highlight the necessity for cross-model communications, where edge devices from one service provider can work with edge servers from another, as illustrated in Fig.\,\ref{fig:intro}. 
However, neural network optimization is significantly influenced by network architectures, weight initialization, training hyperparameters, and other stochastic elements in training processes~\cite{bansal2021revisiting}. Consequently, the independently optimized neural networks result in incoherent feature spaces~\cite{lenc2015understanding, kornblith2019similarity} and incur formidable obstacles to cross-model communication among different edge AI systems. For instance, the features encoded by the VGG-based encoder cannot be recognized and utilized by ResNet-based neural networks in edge servers from another service provider. Despite sharing the same neural network backbone, the features encoded from distinct training processes are not mutually identifiable by external systems. 
{These challenges motivate us to propose efficient approaches to align the features encoded by diverse task-oriented communication systems. Through feature alignment, we can develop model-agnostic task-oriented communication systems capable of enabling cross-model inference in real-time CV applications.}

\subsection{Contributions}
In this work, we investigate cross-model task-oriented communication in edge AI systems. To enable cross-model inference, we formulate a feature alignment problem and further propose server-based and on-device methods of alignment according to the intrinsic traits of visual representations encoded by neural networks. 
\begin{itemize}
    \item We formulate the feature alignment problem for cross-model edge inference between independent task-oriented communication systems. By sharing the anchor data across different systems, we align feature spaces to enable cross-model edge inference. Then, we propose server-based and on-device feature alignment.
    \item For server-based alignment, we capitalize on the linear invariance of visual representations and propose to align two distinct feature spaces by estimating linear transformation using the encoded features of anchor data, which does not introduce any computational and memory overhead to edge devices. 
    \item By leveraging the angle-preserving property of visual feature space, we further propose to transmit the relative representations based on their angle information in task-oriented communications for on-device alignment. The relative representations can be directly recognized by other task-oriented communication systems, eliminating any additional latency in feature alignment processes.
    \item We conduct extensive experiments using various representative neural network architectures for edge inference and computer vision benchmarks. The experimental results demonstrate the effectiveness of the proposed methods for cross-model task-oriented communication. Furthermore, runtime, memory, and computation overhead analysis further confirm the applicability of feature alignment in real-time scenarios.
\end{itemize}

 The rest of the paper is organized as follows. Section~\ref{sec: model} introduces the system model of cross-model task-oriented communications and describes the formulation of feature alignment. Section~\ref{sec: server-based} and Section~\ref{sec: on-device} propose feature alignment methods for server-based and on-device scenarios, respectively. In Section~\ref{sec: exp}, we provide extensive simulation results to evaluate the performance and effectiveness of the proposed feature alignment methods for cross-model edge inference. Finally, Section~\ref{sec: conc} concludes the paper.

Throughout this paper, we denote random variables and their realizations as capital letters (e.g., $X$) and lowercase letters (e.g., $\x$), respectively. Furthermore, the subscripts of random variables and their realizations indicate the belonging of the trained learning system (e.g., $Z_1$ comes from the learning system indexed by $1$). The statistical expectation is denoted as $\bE[\cdot]$, the mutual information between $X$ and $Y$ is denoted as $I(X;Y)$, and the Kullback-Leibler (KL) divergence between two distributions $p(\x)$ and $p(\y)$ is denoted as $\KL(p(\x)\| p(\y))$. We further denote the Gaussian distribution with mean $\boldsymbol{\mu}$ and covariance matrix $\Sigma$ as $N(\boldsymbol{\mu}, \mathbf{\Sigma})$ and use $\mathbf{I}$ to represent the identity matrix.

%% file: secs/3_system.tex
\section{System Model and Problem Formulation}\label{sec: model}
In this section, we introduce probabilistic modeling and the cross-model inference for task-oriented communication. Then, the feature alignment problem is formulated for model-agnostic and cross-model edge inference.
\subsection{Probabilistic Modeling for Task-Oriented Communications}
We consider a point-to-point task-oriented communication system for edge inference, where the neural networks are employed at both the local device and the edge server. We define a data source that generates the data sample $\x$ and corresponding target $\y$ with a joint distribution $p(\x, \y)$. The dataset $\{\x^{(i)} \y^{(i)} \}_{i=1}^n$ comprises $n$ independent and identically distributed (i.i.d.) data samples and targets. At the local device side, the on-device network encodes the input data $\x$ into a feature vector $\z \in \mathcal{Z}$ using a probabilistic encoder $p_{\bphi}(\z|\x)$ parameterized by the trainable parameters $\bphi$. We denote the dimension of the feature vector $\z$ as $d$. 
The encoded feature is transmitted through an additive white Gaussian noise (AWGN) channel, with noise vector $\bepsilon \sim N(\boldsymbol{0}, \sigma^2 \boldsymbol{I})$\footnote{Although we assume AWGN channels for simplicity and consistency with other research on task-oriented communications, this model can be extended to other channel models such as fading channel and MIMO channel as long as the transmitted features can be formulated within the optimization.}. The noisy feature received at the edge server, denoted as $\tz$, is expressed by
\begin{align}
    \tz & = \z + \bepsilon.
\end{align}
The effect of the noisy channel is captured by the conditional distribution $p(\tz|\z)$, which is independent of $\bphi$. This encoding process forms a Markov chain as,
\begin{align}
		Y&\longleftrightarrow X \stackrel{\bphi}{\longleftrightarrow} Z \longleftrightarrow  \Tilde{Z} .\label{eq:markov_chain}
\end{align}
Theoretically, the optimal inference model of the encoder $p_{\bphi}(\z|\x)$ and the channel model $p(\tz|\z)$ is determined by the posterior $p_{\bphi}(\y|\tz)$. However, due to the intractability of the high-dimensional integrals in the posterior computation, we replace the optimal inference model with a variational approximation $p_{\btheta}(\y|\tz)$ parameterized by parameters $\btheta$. At the receiver, the inference model $p_{\btheta}(\y|\tz)$ receives the corrupted features $\hz$ and performs inference to output the result $\hy$.

For training the models in task-oriented communication, a specific channel model shall be considered and the entire models are end-to-end optimized with an objective function. Although there exist diverse objectives for various targets, a fundamental objective is to maximize the task-relevant information within the transmitted features. This problem is typically formulated by optimizing the mutual information between the transmitted feature $\Tilde{Z}$ and the target $Y$, which can be expressed as,  
\begin{align}
     I(\Tilde{Z}; Y)  =& \bE_{p(\x,\y)p_{\bphi}(\tz|\x)}[\log p_{\bphi}(\y|\tz)] + H(Y)\\
     =& \underbrace{\bE_{p(\x,\y)p_{\bphi}(\tz|\x)}[\log p_{\btheta}(\y|\tz)] }_{\mathcal{L}(\bphi, \btheta)} + H(Y)\notag \\
     &+ \bE_{p_{\bphi}(\tz)}\{\underbrace{\bE_{p_{\bphi}(\y|\tz)}[\log \frac{p_{\bphi}(\y|\tz)}{p_{\btheta}(\y|\tz)}]}_{\KL(p_{\bphi}(\y|\tz) \| p_{\btheta}(\y|\tz) \geq 0 } \} \\
     &\geq \mathcal{L}(\bphi, \btheta).
\end{align}
In particular, $H(Y)$ is a constant term and $\KL(p_{\bphi}(\y|\tz) \| p_{\btheta}(\y|\tz)$ is a non-negative term. Thus, $\mathcal{L}(\bphi, \btheta)$ is utilized as a tractable lower bound for $I(\Tilde{Z}; Y)$, which serves as a computable objective for the end-to-end optimization.

\begin{figure}[t]
		\centering
		\includegraphics[width=8.5cm]{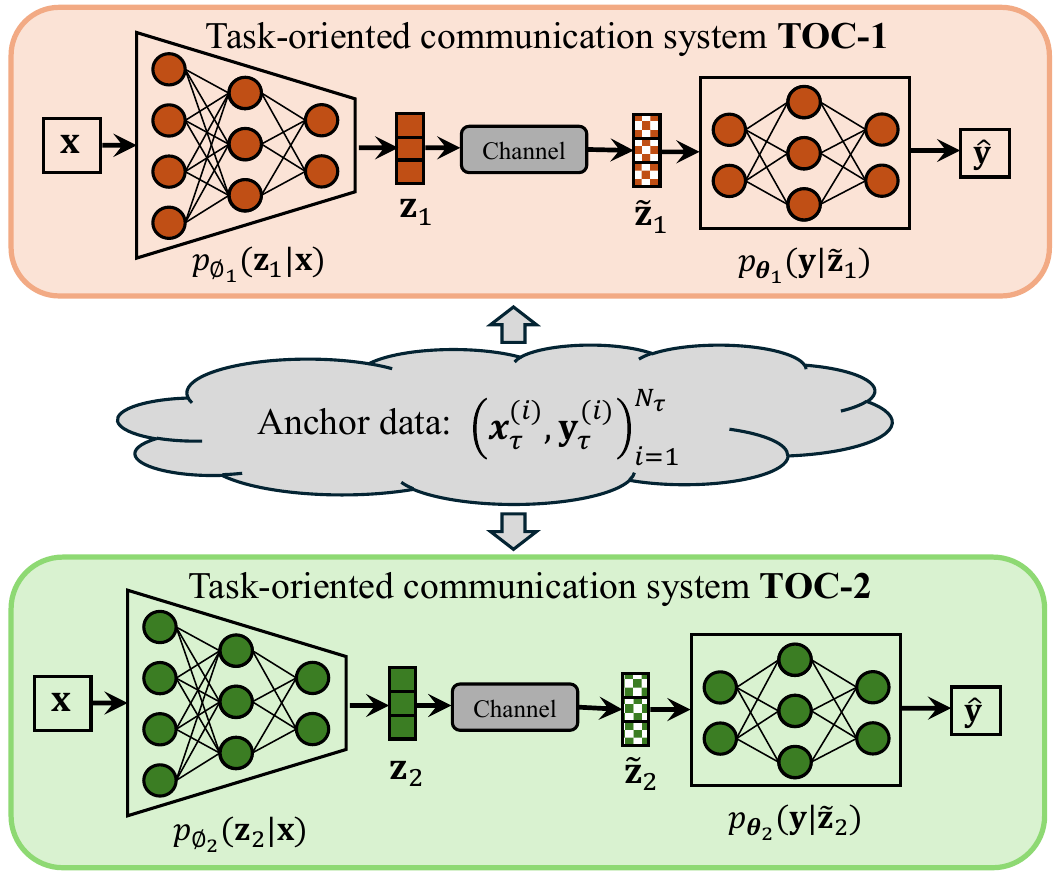}
		\caption{The independent task-oriented communication models, denoted as TOC-1 and TOC-2, and the anchor data shared between them. The parameters $(\bphi_1, \btheta_1)$ and $(\bphi_2, \btheta_2)$ for each task-oriented communication model are trained with different architectures and independent training processes. Anchor data works as a common knowledge base and all the task-oriented communication models can recognize its data samples and data index.
		}
  \label{fig:system_1}
\end{figure}
\subsection{Cross-Model Task-Oriented Communications}
\begin{figure}[t]
		\centering
		\includegraphics[width=6.2cm]{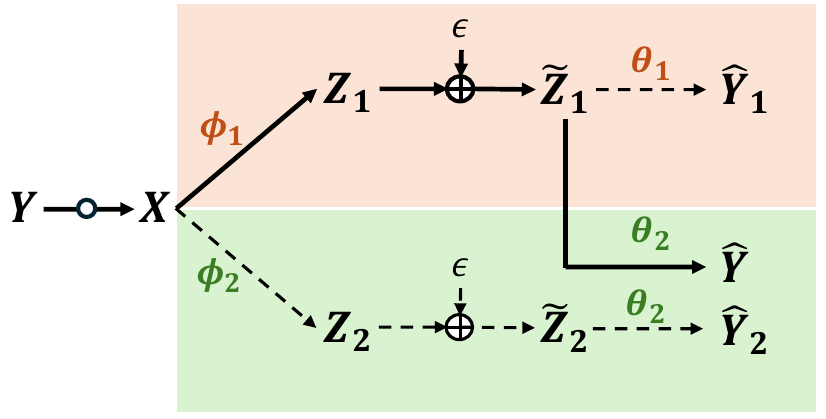}
		\caption{Probabilistic modeling for cross-model task-oriented communication.
		}
  \label{fig: prob_model}
\end{figure}

As illustrated in Fig.\,\ref{fig:system_1}, we investigate the cross-model communication between independent edge inference systems. To solve the problem, we first introduce the \emph{independent task-oriented communication model} and the concept of \emph{anchor data}. 
\begin{itemize}
    \item \textbf{Independent Task-Oriented Communication Models} have different network architectures and are trained independently. The neural network structures and stochastic elements during training, including random weight initialization, data shuffling, and training hyperparameters, can influence the learned parameters. For simplicity, the two independent models are denoted as \emph{TOC-1} and \emph{TOC-2}, and their parameters and encoded features are represented as $(\bphi_1, \btheta_1)$ and $(\bphi_2, \btheta_2)$, $Z_1$ and $Z_2$, respectively.
    \item \textbf{Anchor Data:} Among independent task-oriented communication systems,  there exists a set of shared data points $(\x_\tau^{(i)}, \y_\tau^{(i)})_{i=1}^{n_\tau}$ with a size of $n_{\tau}$, called \emph{anchor data} and denoted as
  \begin{align}
     \mathbf{X}_{\tau} &= [\x_{\tau}^{(1)}, \x_{\tau}^{(2)}, \dots, \x_{\tau}^{(n_\tau)}].
 \end{align}
The anchor data samples, along with their respective order, are identifiable by all task-oriented communication systems. Specifically, the features of anchor data $\x$ encoded by the TOC-$k$ system are denoted as $\z_{\tau, k}$, which are organized in matrix form as
 \begin{align}
     \bZ_{\tau, k} &= [\z_{\tau,k}^{(1)}, \z_{\tau, k}^{(2)}, \dots, \z_{\tau, k}^{(n_\tau)}].
 \end{align}
The incorporation of anchor data aligns with contemporary communication standards and the fundamental concept of a \emph{knowledge base}~\cite{luo2022semantic, zhang2022deep} in semantic communications. It plays a pivotal role in harmonizing feature spaces across diverse task-oriented communication systems.
\end{itemize}
The objectives $\mathcal{L}(\bphi_1, \btheta_1)$ and $\mathcal{L}(\bphi_2, \btheta_2)$ for TOC-$1$ and TOC-$2$ are the lower bounds of task-relevant information $I(\Tilde{Z_1}; Y)$ and $I(\Tilde{Z_2}; Y)$. That is, 
\begin{align}
      I(\Tilde{Z}_1; Y) &\geq \mathcal{L} (\bphi_1, \btheta_1)= \mathbb{E}_{p(\x,\y)p_{\bphi_1}(\tz_1|\x)} [\log p_{\btheta_1}(\y|\tz_1)],\notag\\
     I(\Tilde{Z}_2; Y) &\geq \mathcal{L} (\bphi_2, \btheta_2)= \mathbb{E}_{p(\x,\y)p_{\bphi_2}(\tz_2|\x)} [\log p_{\btheta_2}(\y|\tz_2)].\notag
\end{align}
Note that trained parameters $(\bphi_1, \btheta_1)$ and $ (\bphi_2, \btheta_2)$ for two task-oriented communication models are optimized independently using $\mathcal{L}(\bphi_1, \btheta_1)$ and $\mathcal{L}(\bphi_2, \btheta_2)$. Thus, the distributions of the features $\Tilde{Z}_1$ and $\Tilde{Z}_2$ are incoherent and $I(\Tilde{Z}_1; Y) \not = I(\Tilde{Z_2}; Y)$.
Now we consider the communication between TOC-$1$ and TOC-$2$ models, where the edge server performs inference by leveraging the features encoded by the local device from another system. Without loss of generality, we concentrate on the edge inference process from TOC-1 to the TOC-2, i.e., performing inference with $p_{\bphi_1}(\tz_1|\x)$ and $p_{\btheta_2}(\y|\tz_1)$, as depicted in Fig.~\ref{fig: prob_model}.

Similar to the single task-oriented communication system, the cross-model scheme also aims to preserve sufficient task-relevant information by maximizing the mutual information between the underlined target $Y$ and the observed variables $\Tilde{Z}_1$ as well as the anchor data $\mathbf{X}_{\tau}$. It can be expressed as,
\begin{align}
   I(Y;\Tilde{Z}_1, \mathbf{X}_{\tau})=& I(Y;\Tilde{Z}_1|\mathbf{X}_{\tau})+ \underbrace{I(Y;\mathbf{X}_{\tau})}_{=0}\notag\\ 
   =& I(Y;\Tilde{Z}_1|\mathbf{X}_{\tau}),
\end{align}
where $I(Y;\mathbf{X}_{\tau})=0$ because $Y$ is independent of anchor data $\mathbf{X}_{\tau}$. Therefore, the objective of cross-model task-oriented communications is to maximize the lower bound of $I(Y;\Tilde{Z}_1|\mathbf{X}_{\tau})$. However, unlike maximizing $I(Y; \Tilde{Z}_1)$ where parameters $(\bphi_1, \btheta_1)$ are trained jointly to maximize the lower bound $\mathcal{L}(\bphi_1, \btheta_1)$, the parameters $\bphi_1$ and $\btheta_2$ are optimized separately through different training processes.  
Moreover, since $\bphi_1$ and $\btheta_2$ are associated with different encoded feature spaces $\mathcal{Z}_1$ and $\mathcal{Z}_2$, the features $\tz_1$ transmitted from encoder $p_{\bphi_1}(\tz_1|\x)$ cannot be recognized and utilized by the server-based network with parameters $\btheta_2$. This problem motivates us to investigate the feature alignment for task-oriented communication.

\subsection{Problem Formulation for Feature Alignment}\label{subsec:problem_feature_alignment}
To enable cross-model task-oriented communication, we propose two approaches to align the features, named \emph{server-based} and \emph{on-device} alignment, which are detailed below.

\subsubsection{Server-Based Alignment}\label{subsubsec:server-based}
In the server-based scenario, the feature alignment is implemented in the edge server before the feature transmission. 
The basic idea of the server-based feature alignment is to rapidly estimate the transformation $T(\cdot)$ between encoded anchor features $\mathbf{Z}_{\tau, 1}$ and $\mathbf{Z}_{\tau, 2}$ and then utilize the aligned features $T(\tilde{\z}_1)$ in the following inference.
In particular, server-based feature alignment frameworks are composed of the following two phases.
\begin{itemize}
    \item \textbf{Alignment Phase:} Before performing edge inference, the local device first transmits the features of anchor data $\mathbf{Z}_{\tau, 1}$ encoded by the on-device network $p_{\bphi_1}(\z_1|\x)$. After the noisy channel $p(\tz|\z)$, the transmitted anchor features $\Tilde{\mathbf{Z}}_{\tau, 1}$ can be expressed by 
    \begin{align}
         \Tilde{\mathbf{Z}}_{\tau, 1} &= \mathbf{Z}_{\tau, 1} + \sigma^2\mathbf{\Upsilon},
    \end{align}
    where $\mathbf{\Upsilon} = [\bepsilon^{(1)}, \bepsilon^{(2)}, \dots, \bepsilon^{(n_{\tau})}]$ is the matrix of channel noise. After receiving $\Tilde{\mathbf{Z}}_{\tau, 1}$, the edge server will estimate the transformation $T(\cdot): \mathcal{Z}_1 \to \mathcal{Z}_2$ by leveraging received anchor features $\Tilde{\mathbf{Z}}_{\tau, 1}$ and $\Tilde{\mathbf{Z}}_{\tau, 2}$. 
    \item \textbf{Inference Phase:} During the inference phase, the edge server $p_{\btheta_2}(\tz_1| \x)$ receives the features $\tz_1$ from the edge device and compute the transformed feature $T(\tz_1)$. The server-based network $p_{\btheta_2}(\y| T(\tz_1))$ to output the inference result $\hat{\y}$ according to the transformed features $T(\tz_1)$.
\end{itemize}
The server-based alignment maximizes the $I(Y;\Tilde{Z}_1|\mathbf{X}_{\tau})$ depending on the optimal estimation of transformation $T(\cdot)$ given $\Tilde{\bZ}_{\tau, 1}$ and $\Tilde{\bZ}_{\tau, 2}$. Thus, estimation of optimal $T(\cdot)$ is equivalent to maximize one lower bound of $I(Y;\Tilde{Z}_1|\mathbf{X}_{\tau})$ given parameters $\bphi_1$ and $\btheta_2$,
 \begin{align}
     \mathcal{L}_{\text{OS}} ( T, \bphi_1, \btheta_2) = \mathbb{E}_{p(\x,\y)p_{\bphi_1}(\tz_1|\x)} [\log p_{\btheta_2}(\y|T(\tz_1))].
 \end{align}
 \subsubsection{On-Device Alignment} In contrast to the server-based alignment approach, the on-device alignment method eliminates the need for transmitting $\mathbf{Z}_{\tau, 1}$ and estimating $T(\cdot)$, thereby avoiding additional time overhead. To enable seamless cross-model task-oriented communication without introducing any additional latency, on-device alignment is integrated into the encoding stages at the edge devices, aligning the features with the anchor data. Subsequently, the aligned features are sent to the edge server for inference without requiring further processes.
 
 Therefore, in the on-device scenario, the data $\x$ is encoded into features and then aligned with $\mathbf{X}_{\tau}$, which can be formulated by a probabilistic encoding $p_{\bphi_1}(\z_1|\x, \mathbf{X}_{\tau})$. Then, the aligned features $\z_1$ are transmitted to the server-based network $p_{\btheta_2}(\y|\tz_2)$ to perform the cooperative inference. Therefore, the on-device alignment aims to preserve the task-relevant information $I(Y;\Tilde{Z}_1|\mathbf{X}_{\tau})$ by maximizing the following lower bound $\mathcal{L}(\bphi_1, \btheta_2)$,
 \begin{align}
     \mathcal{L}_{\text{ZS}} ( \bphi_1, \btheta_2) = \mathbb{E}_{p(\x,\y)p_{\bphi_1}(\tz_1|\x, \mathbf{X}_{\tau})} [\log p_{\btheta_2}(\y|\tz_1)].
 \end{align}

%% file: secs/4_method_1.tex
\section{Server-Based Feature Alignment}\label{sec: server-based}
In this section, we propose the server-based feature alignment for cross-model task-oriented communication, which leverages the features of anchor data to estimate the transformation between two distinct feature spaces. 
Based on the linear invariance and identifiability within neural network representations, we present theoretical guarantees and develop efficient server-based feature alignment methods through the estimation of linear transformations.
\subsection{Linear Invariance of Semantic Features} 
Recent empirical studies found the invariance of feature spaces up to linear transformations for various neural features, especially for visual representations~\cite{lenc2015understanding, kornblith2019similarity, moschella2022relative}. The concept of linear invariance is further supported by the theoretical justifications for a variety of linearly identifiable models based on the classical statistical methods including the nonlinear independent component analysis (ICA)~\cite{khemakhem2020variational} and energy-based model~\cite{roeder2021linear}.
We assume equality between semantic feature spaces $\mathcal{\hat{Z}}_1$ and $\mathcal{Z}_2$ up to a linear transformation $\mathbf{M}$ as follows.
\begin{assumption}\label{ass-1}
Given a data point $\x$, the features $\z_1$ and $\z_2$ are encoded by the neural network encoders $p_{\bphi_1}(\z_1|\x)$ and $p_{\bphi_2}(\z_2|\x)$. The relationship between features $\mathbf{z}_1$ and $\mathbf{z}_2$ is defined by an injective linear transformation $\mathbf{M}\in \mathbb{R}^{d\times d}$ as
 \begin{align}
     \z_2 &= \mathbf{M}\z_1.
 \end{align}  
\end{assumption}

This linear invariance allows to adopt M as the transformation to align the received features $\tz_1$. Thus, the aligned features $T(\tz)$ can be expressed by
\begin{align}
    T(\tz_1) & = \M \tz_1 = \z_2 + \M\bepsilon.
\end{align}
Obviously, the aligned features $T(\tz_1)$ consist of features $\z_2$ and transformed noise $\M\bepsilon$, which only differ from the target features $\tz_2$ by the noise component as $\tz_2 = \z_2 + \bepsilon$. The difference between $T(\tz_1)$ and $\tz_2$ determines the performance of the cross-model edge inference with the alignment $\M$. As presented in Section~\ref{subsec:problem_feature_alignment}, the effectiveness of feature alignment $T=\M$ can be expressed by the obtained lower bound of task-relevant information,
 \begin{align}
     \mathcal{L}_{\text{OS}} ( \mathbf{M}, \bphi_1, \btheta_2) = \mathbb{E}_{p(\x,\y)p_{\bphi_1}(\tz_1|\x)} [\log p_{\btheta_2}(\y|\mathbf{M}\tz_1)].
 \end{align}
The key for investigating the value of $\mathcal{L}_{\text{OS}} ( \mathbf{M}, \bphi_1, \btheta_2)$ is to interpret the effect of difference between $\M\tz_1$ and $\tz_2$ on the server-based network $p_{\btheta_2}(\y|\tz_2)$. The probabilistic network $p_{\btheta_2}(\y|\tz_2)$ is formulated by the neural network with parameters $\btheta_2$ such that the output logits of the neural network present the $\log p_{\btheta_2}(\y|\tz_2)$ for all $\y \in \mathcal{Y}$. To analyze the neural networks, we assume Lipschitz continuity for neural networks in edge servers, which is widely adopted in deep learning analysis such as robustness against adversarial attacks, smooth neural representation learning, and model generalization~\cite{virmaux2018lipschitz}.
\begin{definition}(Lipschitz Continuity).
Let $f_{\btheta}: \mathcal{Z} \to \mathcal{R}$ be a neural network with parameters $\btheta$. The network is Lipschitz continuous if there exists a non-negative constant $\rho$ such that for any $\z', \z'' \in \mathcal{Z}$,
\begin{align}
    \|f_{\btheta}(\z') - f_{\btheta}(\z'') \|_p \leq \rho\| \z' -\z''\|_p,
\end{align}
where $\| \cdot \|_p$ denotes the $p$-norm. The non-negative $\rho$ is called the Lipschitz constant.
\end{definition}
Assuming Lipschitz continuity for the server-based network $p_{\btheta_2}(\y|\tz_2)$, we can obtain the gap between $\mathcal{L}_{\text{OS}} ( \mathbf{M}, \bphi_1, \btheta_2)$ and the lower bound of TOC-2 system $\mathcal{L}(\bphi_2, \btheta_2)$, illustrated as the following proposition.
\begin{proposition}
\label{prop:linear}
Suppose the server-based decoder $p_{\btheta_2}(\y|\tz_2)$ is $\rho$-Lipschitz smooth and the encoded feature space ${\mathcal{Z}}_1$ and ${\mathcal{Z}}_2$ are invariant up to a linear transformation $\M$, where $\M$ is a full rank matrix. If $\mathbf{M}$ is utilized in the server-based feature alignment scenarios, the difference between the corresponding lower bound $\mathcal{L}_{\text{OS}} (\M, \bphi_1, \btheta_2)$ and $\mathcal{L}(\bphi_2, \btheta_2)$ can be upper-bounded by,
\begin{align}
    | \mathcal{L}_{\text{OS}} ( \M, \bphi_1, \btheta_2) - \mathcal{L}(\bphi_2, \btheta_2)| \leq  \sigma \sqrt{2\log d} \| \M -\mathbf{I} \|_{\infty},
\end{align}
\begin{proof}
 The proof is provided in Appendix~\ref{appendix a}.
\end{proof}
\end{proposition}
Proposition~\ref{prop:linear} shows that the utility gap between $\mathcal{L}_{OS}(\M, \bphi_1, \btheta_2)$ and $\mathcal{L}(\bphi_2, \btheta_2)$ is determined by $\| \M -\mathbf{I} \|_{\infty}$, the dimensionality of encoded features $d$, channel noise level $\sigma$, and Lipschitz constant $\rho$. Note $\| \M -\mathbf{I} \|_{\infty}$ is intrinsically limited when the encoded features are normalized to meet the transmission power constraint. 

\subsection{Linear Transformation Estimation} 
\begin{figure}[t]
		\centering
		\includegraphics[width=8.8cm]{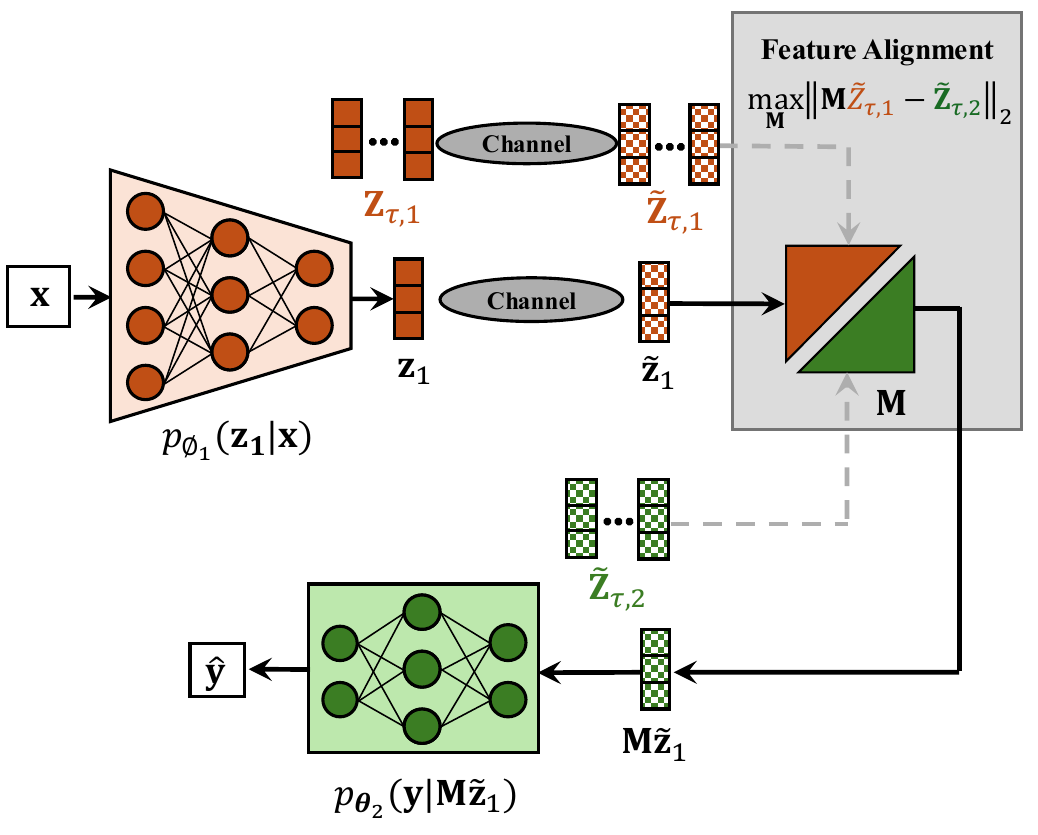}
		\caption{The server-based feature alignment using single-layer MLP for cross-model task-oriented communications.
		}\label{fig: ill-server-based}
\end{figure}
The previous section shows the linear transformation $\M$ between the encoded feature spaces $\mathcal{Z}_1$ and $\mathcal{Z}_2$, which can be utilized to achieve feature alignment for cross-model task-oriented communications. 
Following the server-based scenario presented in Section~\ref{subsubsec:server-based}, the edge server needs to estimate $\M$ based on the transmitted anchor features $\Tilde{\mathbf{Z}}_{\tau,1}$ and $\Tilde{\mathbf{Z}}_{\tau,2}$. The transformation can be expressed as,
\begin{align}
    \Tilde{\bZ}_{\tau, 2} &= \M \bZ_{\tau, 1} + \mathbf{\Upsilon}\\
    &= \M(\Tilde{\bZ}_{\tau, 1} -\mathbf{\Upsilon}) + \mathbf{\Upsilon}.\label{eq: mmse_helper}
\end{align} To estimate the linear transformation $\M$, we propose three methods including the learning-based estimator, the least squares (LS) estimator, and the minimum mean-square error (MMSE) estimator.
\subsubsection{Learning-Based Estimator}
It uses a single-layer multilayer perception (MLP) to approximate $\M$ by minimizing the distance of anchor features $\tz^{(j)}_{\tau, 2}$ and the transformed features $\M \tz^{(j)}_{\tau,1}$. The learning-based feature alignment is illustrated in Fig.\,\ref{fig: ill-server-based} and the training objective function can be expressed as,
\begin{align}\label{eq: MLP}
    \hMML & = \argmax\limits_{\M} \frac{1}{n_{\tau}} \sum\limits_{j=1}^{n_{\tau}}\|\tz^{(j)}_{\tau, 2} - \M \tz^{(j)}_{\tau,1} \|_2.
\end{align}
\subsubsection{LS Estimator} Under the assumption of linear invariance, the estimation of $\M$ can be considered as a typical linear regression problem. 
The LS estimator $\hMLS$ can be utilized to minimize the sum of the squared
distance between $\tz^{(j)}_{\tau, 2}$ and $\M \tz^{(j)}_{\tau,1}$ for total $n_{\tau}$ anchor data as
\begin{align}\label{eq: LS}
\hMLS &=  \Tilde{\mathbf{Z}}_{\tau,2}\mathbf{Z}^T_{\tau,1}(\mathbf{Z}_{\tau,1}\mathbf{Z}^T_{\tau,1})^{-1}.
\end{align}

\subsubsection{MMSE Estimator} 
By assuming the prior distribution that $\M$ comprises independent entries with zero-mean unit variance, the MMSE estimator can be utilized to estimate the transformation as
\begin{align}
    \hMSE &= \Tilde{\mathbf{Z}}_{\tau,2} \mathbf{A}^*
\end{align}
where $\mathbf{A}$ is obtained by
\begin{align}
    \mathbf{A}^* &= \argmin\limits_{\mathbf{A}} \mathbb{E} [\|\M - \Tilde{\mathbf{Z}}_{\tau,2} \mathbf{A}\|^2_F.
\end{align}
Furthermore, we assume the prior distribution of each entry of $\M$ to be i.i.d. Gaussian variable $N(0, 1/\sqrt{d})$. 
Using \eqref{eq: mmse_helper} to compute the MSE, the error can be written as 
\begin{align}
    \textrm{MSE} & = \mathbb{E} [\|\M -  (\M(\Tilde{\bZ}_{\tau, 1} -\Upsilon) + \Upsilon)\mathbf{A}\|^2_F].
\end{align}
The optimal $\mathbf{A}^*$ can be derived by $\partial \varepsilon/ \partial \mathbf{A}=0$. Then, we can obtain a linear estimator that minimizes the MSE of $\M$, which can be expressed as,
\begin{align}\label{eq: MMSE}
\hMSE & = \Tilde{\bZ}_{\tau, 2}(\Tilde{\bZ}_{\tau, 1}^T \Tilde{\bZ}_{\tau, 1} + 2n_{\tau}\sigma^2 \mathbf{I})^{-1}\Tilde{\bZ}_{\tau, 1}^T.
\end{align}
The detailed derivation of $\mathbf{A}$ is provided in Appendix~\ref{appendix b}.
After estimating the linear transformation $\M$, the proposed server-based feature alignment for cross-model task-oriented communications is summarized in Algorithm~\ref{algo-server-based}.
\begin{algorithm}[t]
\small
\caption{Cross-Model Inference with Server-Based Feature Alignment}
\begin{algorithmic}[1]
\label{algo-server-based}
\REQUIRE Anchor data $(\x_\tau^{(i)}, \y_\tau^{(i)})_{i=1}^{n_{\tau}}$, optimized parameters $\bphi_1$, $\btheta_2$, channel variance $\sigma^2$, and the noisy anchor features $\Tilde{\bZ}_{\tau, 2}$ stored in the edge server of TOC-2.
\\ \underline{\# Alignment Phase}
\STATE The on-device encoder $p_{\bphi_1}(\z|\x)$ computes the features ${\bZ}_{\tau, 1}$ of anchor data $(\x_\tau^{(i)})_{i=1}^{n_{\tau}}$ .
\STATE The device sends ${\bZ}_{\tau, 1}$ across AWGN channels
\STATE The edge server receives noisy anchor features $\Tilde{\bZ}_{\tau, 1}$.
\STATE \textbf{Option \Rmnum{1}} (Learning-based Estimation):
\STATE \hspace{4mm} Initialize the single-layer MLP for $\M$
\STATE \hspace{4mm} Optimize $\M$ using the loss function \eqref{eq: MLP}
\STATE \textbf{Option \Rmnum{2}} (LS Estimation):
\STATE \hspace{4mm} Compute $\M$ based on \eqref{eq: LS}
\STATE \textbf{Option \Rmnum{3}} (MMSE Estimation):
\STATE \hspace{4mm} Compute $\M$ based on \eqref{eq: MMSE}
\\ \underline{\# Inference Phase}
\STATE The on-device encoder $p_{\bphi_1}(\z|\x)$ computes the features $\z_1$ 
\STATE The device sends $\z_1$ across AWGN channels
\STATE The edge server receives noisy features $\tz_1$
\STATE The edge server aligns $\tz_1$ using estimated transformation $\M$
\STATE The server-based $p_{\btheta_2}(\y|\tz_2)$ leverages $T(\tz_1)$ outputs target $\hat{\y}$
\end{algorithmic}
\end{algorithm}

%% file: secs/4_method_2.tex
\section{On-device Feature Alignment}\label{sec: on-device}
In this section, we propose the on-device feature alignment where the local device aligns the encoded features using the anchor data within the end-to-end training. With the proposed on-device alignment, the on-device encoder encodes the data sample into aligned features that can be identified and utilized by the edge server from other deployed edge servers. 
\subsection{Relative Representations}
In the on-device feature alignment scenarios, we aim to design the on-device encoder that encodes the features that other edge servers can identify. The direct solution is to ensure the features encoded by different on-device encoders are equivalent. That is, given the data sample $\x$, the encoded features from $p_{\bphi_1}(\z_1|\x)$ and $p_{\bphi_1}(\z_2|\x)$ are identical, 
\begin{align}
    \z_1 =\z_2,\text{ where } \z_1 \sim p_{\bphi_1}(\z_1|\x,\mathbf{X_{\tau}}),\z_2 \sim p_{\bphi_2}(\z_2|\x,\mathbf{X}_{\tau}). \notag
\end{align}

Intuitively, we expect the independent on-device neural networks output the same feature space regardless of their architectures or any random factors in the training process such as hyperparameters and optimization algorithms. Recent studies empirically demonstrate that the feature space transformation induced by those random factors is \emph{angle-preserving}~\cite{moschella2022relative}.  We introduce this intrinsic behavior for neural feature encoding in the following assumption.
\begin{assumption}\label{ass-2}
    For any two data samples $(\x', \x'') \in \mathcal{X}$ and the corresponding features $(\z_1', \z_1'')$ and $(\z_2', \z_2'')$ encoded by $p_{\bphi_1}(\z_1|\x)$ and $p_{\bphi_1}(\z_2|\x)$, respectively, we have
    \begin{align}
        \angle(\z_1', \z_1'') = \angle(\z_2', \z_2'').
    \end{align}
\end{assumption}
Based on Assumption~\ref{ass-2}, the angles between the features encoded from the same data samples are the same for different neural network encoding methods. Assumption~\ref{ass-2} and Assumption~\ref{ass-1} are intricately related. This is because feature vectors are equivalent up to a linear transformation, and their angles remain consistent if they are $\ell_2$-normalized. Therefore, for any input data $\x$, the angles between $\x$ and a set of anchor data $\{\x_{\tau}^{(i)} \}_{i=1}^{n_\tau}$ are identical for on-device encoders from independent task-oriented communication systems. 

By utilizing the angle-preserving property, we propose to align encoded features with anchor features and then transmit their angle-related information to edge servers. Firstly, we adopt the cosine similarity to represent the angle information between feature vectors, given two encoded feature vectors $\z'$ and $\z''$, the cosine similarity $\Sc(\z', \z'')$ can be expressed as,
\begin{align}
    \Sc(\z', \z'') & = \frac{\z'^T \z''}{\| \z'\|_2\| \z'' \|_2}.
\end{align}
\begin{figure}[t]
		\centering
		\includegraphics[width=8.cm]{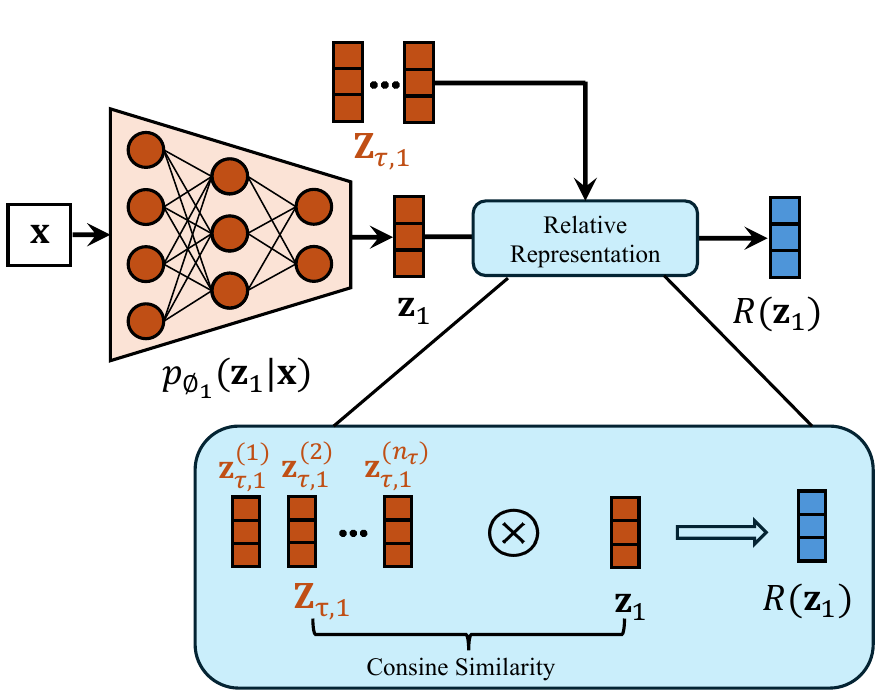}
		\caption{The proposed on-device feature alignment using relative representation encoding.
		}
  \label{fig:on-device}
\end{figure}
As illustrated in Fig.\,\ref{fig:on-device}, the feature vector $\z$ of every sample $\x$ will be represented by the cosine similarity with respect to the encoded anchor features. Given the feature vector $\z$, the on-device encoder $p_{\bphi_k}(\z|\x)$ outputs the relative representation $R(\z; \mathbf{X}_{\tau}, \bphi_i)$ by computing the cosine similarity with anchor features $\bZ_{\tau,k}$ as
\begin{align}\label{eq:relative_repr}
    R(\z; &\mathbf{X}_{\tau}, \bphi_k)\notag \\
    &= [\Sc(\z, \z_{\tau,k}^{(1)}), \Sc(\z, \z_{\tau, k}^{(2)}), \dots, \Sc(\z, \z_{\tau, k}^{(n_\tau)})].
\end{align}
Different from transmitting $\z$, the dimension of $R(\z; \mathbf{X}_{\tau}, \bphi_k)$ is determined by the number of anchor data, $n_{\tau}$, instead of $d$. Thus, the transmission latency will be larger if a larger number of anchor data samples are considered in alignment. 
\subsection{On-Device Feature Alignment with Relative Representations}
Here we present the training algorithm with the on-device feature alignment by utilizing relative representations. 
For the $k$-th task-oriented communication system, the one-device encoder with parameters $\bphi_k$ computes the representative representation $R(\z; \mathbf{X}_{\tau}, \bphi_k)$ and transmit it over noisy channels to the edge server. The edge server receives the noisy representation $\Tilde{R}(\z; \mathbf{X}_{\tau}, \bphi_k)$ and performs the inference using the server-based network with parameters $\btheta_k$. Following the typical training algorithm, the parameters $\bphi_k$ and $\btheta_k$ are end-to-end optimized by maximizing the on-device lower bound,
\begin{align}\label{eq: loss_zs}
    &\Tilde{L}_{\text{ZS}} (\bphi_i, \btheta_i) = \frac{1}{n}\sum\limits_{j=1}^{n}\left[ \log p_{\btheta_i}(\y^{(j)}|\Tilde{R}(\z_i^{(j)}; \mathbf{X}_{\tau}, \bphi_i))\right],
\end{align}
where $\Tilde{R}(\z_i^{(j)}; \mathbf{X}_{\tau}, \bphi_i) = R(\z_i^{(j)}; \mathbf{X}_{\tau}, \bphi_i) + \bepsilon^{(j)}$,
    $\z^{(i)}_{i} \sim p_{\bphi_i}(\z_{i}|\x^{(i)})$, and $\bepsilon^{(j)} \sim N(\boldsymbol{0}, \sigma^2 \boldsymbol{I})$.
The entire training process with the proposed on-device feature alignment is presented in Algorithm~\ref{algo-on-device}. 

Owing to the relative representation $R(\z; \mathbf{X}_{\tau}, \bphi_k)$, the probabilistic encoder $p_{\bphi_k}(\z_k|\x, \X_{\tau})$ outputs identical representation $\z_k$ for different task-oriented communication systems. For TOC-$1$ and TOC-$2$ trained independently with Algorithm~\ref{algo-on-device}, the encoders $p_{\bphi_1}(\z_1|\x)$ and $p_{\bphi_1}(\z_2|\x)$ output identical features for any data point $\x \in \mathcal{X}$. Therefore, the trained server-based networks $p_{\btheta_1}(\y|\tz_1)$ and $p_{\btheta_2}(\y|\tz_2)$ have the same distribution. Thus, the on-device lower bounds satisfy
\begin{align}
    \mathcal{L}_{ZS} (\bphi_1, \btheta_1)= \mathcal{L}_{ZS} (\bphi_2, \btheta_2)=\mathcal{L}_{ZS} (\bphi_1, \btheta_2).\label{eq:zero}
\end{align}
The equality \eqref{eq:zero} shows that the on-device alignment ensures the cross-model task-oriented communications have the same inference performance with the single model.
\begin{algorithm}[t]
\small
\caption{Training with on-device Feature Alignment}
\begin{algorithmic}[1]
\label{algo-on-device}
\REQUIRE $T$ (number of epochs), batch size $N$, channel variance $\delta^2$, anchor data $(\x_\tau^{(i)}, \y_\tau^{(i)})_{i=1}^{n_\tau}$, initialized parameters $\bphi_k$ and $\btheta_k$ for task-oriented communication system TOC-$k$.
\ENSURE The optimized parameters $\bphi_k$ and $\btheta_k$
\WHILE{epoch $t=1$ to $T$}
    \STATE Sample a mini-batch of data samples $\{(\x^{(i)}, \y^{(i)})\}_{i=1}^{N}$.
    \STATE Compute the encoded feature vectors $\{\z^{(i)}\}_{i=1}^{N}$ of data samples $\{\x^{(i)}\}_{i=1}^{N}$.
    \STATE Compute the encoded feature vectors $\{\z_{\tau}^{(i)}\}_{i=1}^{N}$ of the anchor data  $\{\x_\tau^{(i)}\}_{i=1}^{n_\tau}$.
    \STATE Compute the relative representation $\{R(\z^{(i)})\}_{i=1}^{N}$ based on $\{\z_{\tau}^{(i)}\}_{i=1}^{N}$ using~\eqref{eq:relative_repr}.
    \STATE Sample the noise $\{\boldsymbol{\epsilon}^{(i,l)}\}_{l=1}^L \sim N(0, \sigma^2\mathbf{I})$ and estimate the corrupted feature vectors $\{\tz^{(i,l)}\}_{l=1}^L$.
    \STATE Compute the loss $\tilde{\mathcal{L}}_{\textrm{ZS}}(\bphi_k, \btheta_k)$ based on~\eqref{eq: loss_zs}.
    \STATE Update the parameters $\bphi_k$ and $\btheta_k$ through backpropagation.
\ENDWHILE
\end{algorithmic}
\end{algorithm}

%% file: secs/5_exp.tex
\section{Experiments}\label{sec: exp}
In this section, we conduct extensive experiments the proposed server-based and on-device alignment algorithms with image classification benchmarks. Furthermore, we show the performance of the running time, computational complexity, and memory cost\footnote{The source code: https://github.com/SongjieXie/feature-alignment-TOC}.

\subsection{Experimental Setup}
\subsubsection{Datasets} The experiments are conducted on two benchmark datasets, the Street View House Numbers (\emph{SVHN})~\cite{netzer2011reading} and \emph{CIFAR-10}~\cite{krizhevsky2009learning}. The Street View House Numbers (SVHN) dataset comprises real-world images of house numbers sourced from Google Street View. It consists of over 600,000 labeled RGB images of house numbers extracted from more than 200,000 unique addresses. On the other hand, the CIFAR-10 dataset consists of 32×32 color images categorized into 10 classes, containing 50,000 training images (5,000 per class) and 10,000 test images. 
Data augmentation techniques are applied to both datasets including random cropping and horizontal flipping.
\subsubsection{Neural Network Architectures}  As we consider image classification as the benchmark task, the representative neural network architectures are adopted for the on-device encoder and server-based network. Following the previous research, VGG~\cite{simonyan2014very} and ResNet~\cite{he2016deep} are chosen as the backbones, which are then tailored for edge intelligence systems by splitting the networks and presented in Table~\ref{tab: NN-architecture}.

\begin{itemize}
    \item \textbf{VGG-based}: The VGG-based architecture mainly consists of convolutional (Conv) and fully connected neural network layers.
    \item \textbf{ResNet-based}: The ResNet-based architecture mainly consists of resnet blocks and fully connected layers, which have higher capacity but require more dense computation than VGG-based ones.
    \item \textbf{Mixed}: To balance the model capacity and computational resource demand, another mixed architecture is adopted to tailor the neural network to edge intelligence systems by putting more parameters to edge servers. 
\end{itemize}  
\begin{table*}[t]
		\caption{The Network Architectures of On-device Encoders and Server-based Networks for Image Classification}
		\label{tab: NN-architecture}
        \centering
		 	\resizebox{1.0\linewidth}{!}{
		\begin{tabular}{c|llcc|llcc}
			\toprule
			& \multicolumn{4}{c|}{\textbf{On-device Encoder}} & \multicolumn{4}{c}{\textbf{Server-based Network}} \\
            & Layer & Output Dimension & \#Params & Size & Layer & Output Dimension & \#Params & Size\\
			\midrule
			\multirow{4}{*}{\textbf{VGG-based}} & Conv & $64 \times 32 \times 32$ &\multirow{4}{*}{1,571,416} &\multirow{4}{*}{5.99 MB}&Fully-connected $\times$ $3$ & $64$& \multirow{4}{*}{610,240 }&\multirow{4}{*}{2.33 MB}\\
                                                  & Conv $\times$ $4$ & $4 \times 4 \times 4$& & & Conv $\times$ $2$& $512 \times 4 \times 4$& &\\
                                                  & Reshape & $64$& & &Pooling layer& $512 \times 1 \times 1$ &&\\
                                                  & Fully-connected& $16$ & &&Fully-connected& $10$ &&\\
                \midrule
                \multirow{4}{*}{\textbf{ResNet-based}} & Conv  & $64 \times 32 \times 2$& \multirow{4}{*}{4,845,816} & \multirow{4}{*}{18.49 MB}& Fully-connected $\times$ $3$& $64$& \multirow{4}{*}{3,302,208} & \multirow{4}{*}{12.60 MB}\\
                                                  & ResBlock $\times$ $4$ & $4 \times 4 \times 4$&&& ResBlock $\times$ $3$ & $512 \times 4 \times 4$&&\\
                                                  & Reshape & $64$&& &Pooling layer & $512 \times 1 \times 1$&&\\
                                                  & Fully-connected & $16$& &&Fully-connected & $10$&&\\
                \midrule
			\multirow{5}{*}{\textbf{Mixed}} & Conv $\times$ $2$ & $128 \times 16 \times 16$ & \multirow{5}{*}{1,866,840} & \multirow{5}{*}{7.12 MB}& Fully-connected $\times$ $3$& $64$& \multirow{5}{*}{4,754,624} & \multirow{5}{*}{18.14 MB}\\
                & ResBlock& $128 \times 16 \times 16$ &&& Conv & $512 \times 4 \times 4$&&\\
			& Conv $\times$ $3$ & $4 \times 4 \times 4$& && Resblock& $512 \times 4 \times 4$&&\\
                & Reshape & $64$& &&Pooling layer & $512 \times 1 \times 1$&&\\
                & Fully-connected & $16$& &&Fully-connected& $10$&&\\
			\bottomrule
		\end{tabular}}
	\end{table*}
 \begin{figure}
		\centering
		\subfloat[VGG-based]{
			\centering
			\includegraphics[width=0.305\linewidth]{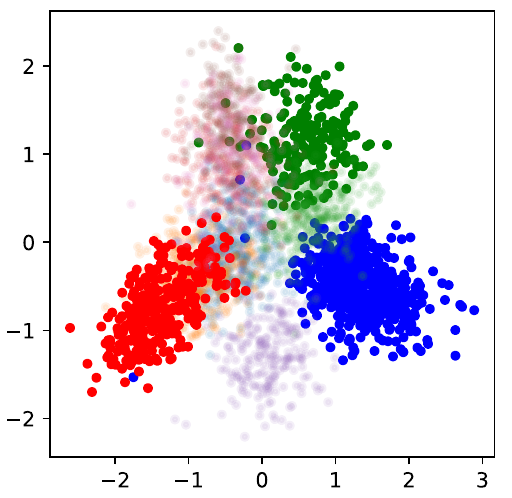}
		}
        \subfloat[ResNet-based]{
			\centering
			\includegraphics[width=0.305\linewidth]{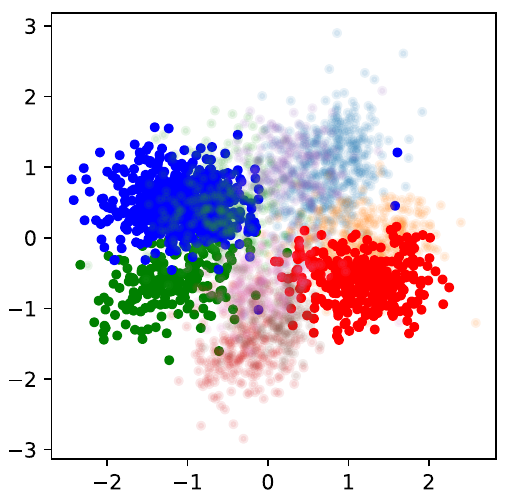}
		}
		\subfloat[Mixed]{
			\centering
			\includegraphics[width=0.3\linewidth]{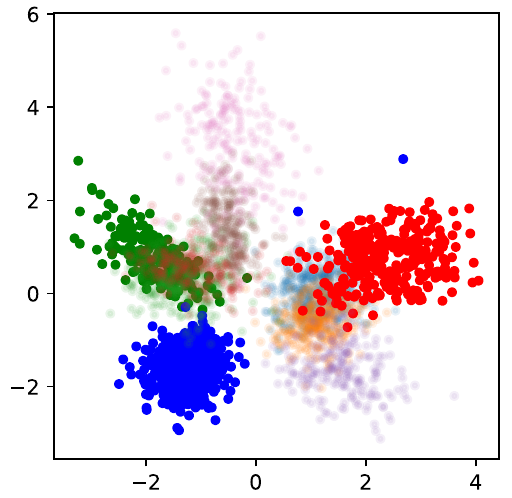}
		}
		\caption{PCA visualization of the feature spaces encoded from the distinct training processes with different neural network architectures: (a) VGG-based, (b) ResNet-based, and (c) Mixed.
		}
		\label{fig:ab-align}
	\end{figure}
 \begin{figure}
		\centering
		\subfloat[VGG-based]{
			\centering
			\includegraphics[width=0.305\linewidth]{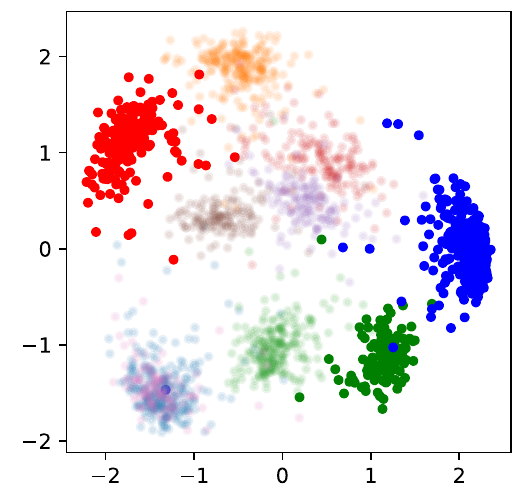}
		}
        \subfloat[ResNet-based]{
			\centering
			\includegraphics[width=0.305\linewidth]{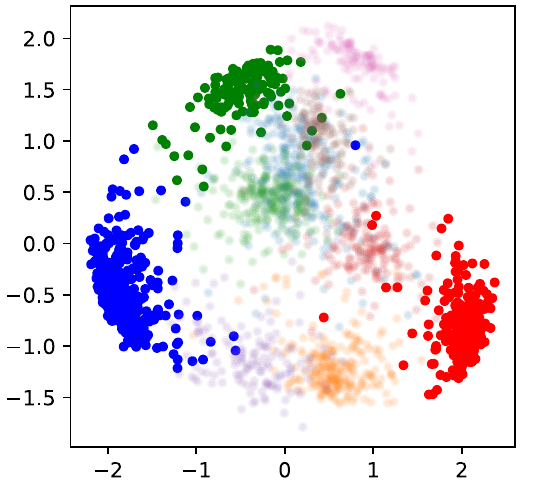}
		}
		\subfloat[Mixed]{
			\centering
			\includegraphics[width=0.3\linewidth]{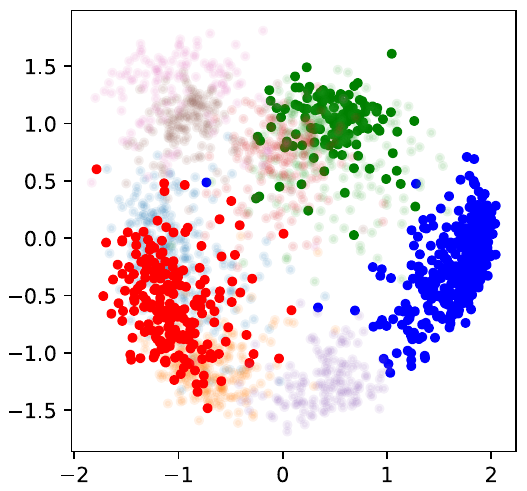}
		}
		\caption{PCA visualization of the relative representations encoded from the distinct training processes with different neural network architectures: (a) VGG-based, (b) ResNet-based, and (c) Mixed.
		}
		\label{fig:ab-rel}
	\end{figure}
\subsubsection{Proposed Methods and Baselines} We focus on the performance of the proposed algorithms for server-based and on-device feature alignment in cross-model task-oriented communications. As presented in Section~\ref{sec: server-based}, the proposed server-based alignment with the LS estimator, the MMSE estimator, and single-layer MLP linear transformation estimation are denoted as \emph{OS-LS}, \emph{OS-MMSE}, and \emph{OS-MLP}, respectively. Besides, we adopt two baselines to evaluate the performance of the proposed server-based alignment methods, denoted as \emph{Non-Alignn} and \emph{FT}, respectively. Non-Alignn is the cross-model task-oriented communications without any feature alignment and FT is the commonly used fine-tuning technique that utilizes multiple-layer neural networks to align the transmitted feature vectors. Furthermore, the on-device alignment is trained independently as Algorithm~\ref{algo-on-device} and is evaluated at the same setting but without additional alignment processes.
\subsubsection{Evaluations} In our experiments, we initially train multiple edge inference systems independently based on the three selected neural network architectures. Subsequently, we interchange the optimized on-device encoders and server-based networks across different systems and perform edge inference using the exchanged encoders and inference networks. 

We consider classification tasks using the SVHN and CIFAR-10 datasets. The standard metric top-1 accuracy is used to evaluate the classification results and a higher accuracy value indicates superior performance in classification.
Note that in edge inference systems, the communication latency is determined by the dimension $d$ of the transmitted feature vector and we consider a fixed dimensionality $d=16$ throughout our experiments. Additionally, we set the number of anchor data samples $n_{\tau}=100$ for the server-based alignment techniques and $n_{\tau}=32$ for the on-device alignment approach.
\subsection{Experimental Results}
 \subsubsection{Feature Space Visualization}
  \begin{figure*}
		\centering
		\subfloat[Non-Align ($\textrm{SNR}=6\textrm{dB}$)]{
    \label{subfig: SVHN-6-nonalign}
			\centering
			\includegraphics[width=0.21\linewidth]{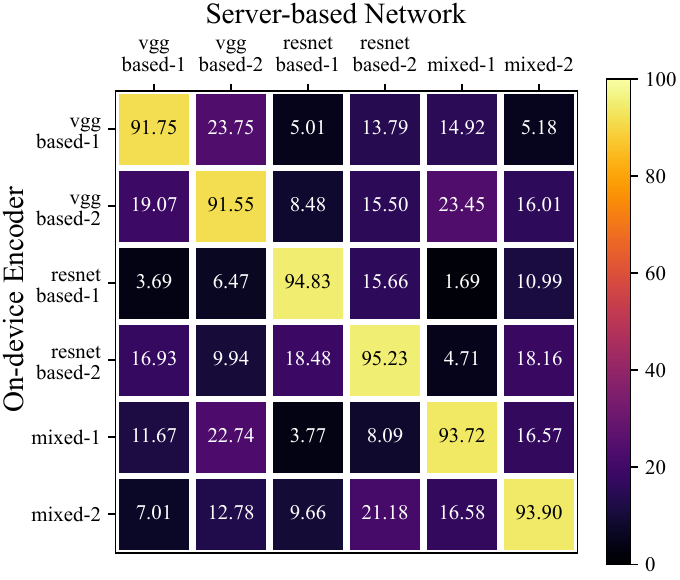}
		}
        \subfloat[FT ($\textrm{SNR}=6\textrm{dB}$)]{
        \label{subfig: SVHN-6-FT}
			\centering
			\includegraphics[width=0.18\linewidth]{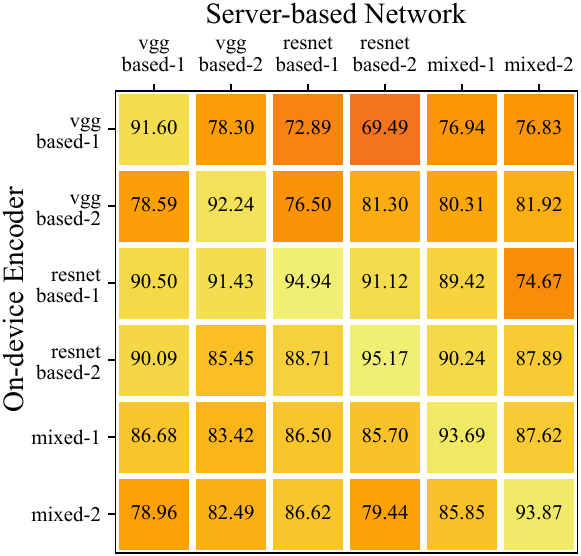}
		}
		\subfloat[OS-LS ($\textrm{SNR}=6\textrm{dB}$)]{
			\centering
			\includegraphics[width=0.18\linewidth]{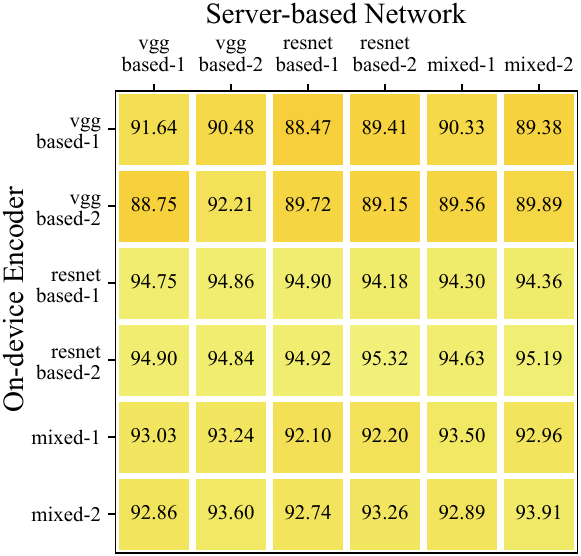}
		}
		\subfloat[OS-MMSE ($\textrm{SNR}=6\textrm{dB}$)]{
			\centering
			\includegraphics[width=0.18\linewidth]{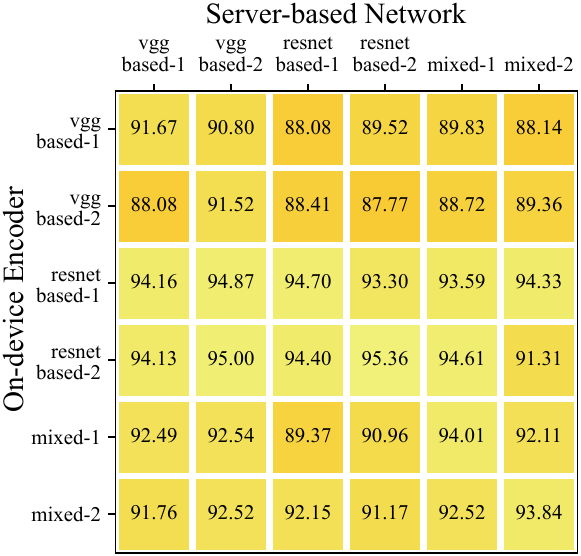}
		}
        \subfloat[OS-MLP ($\textrm{SNR}=6\textrm{dB}$)]{
			\centering
			\includegraphics[width=0.18\linewidth]{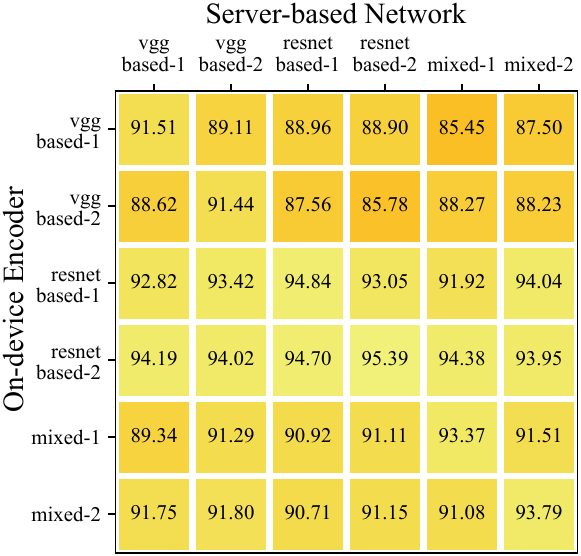}
		}

        \subfloat[Non-Align ($\textrm{SNR}=18\textrm{dB}$)]{
        \label{subfig: SVHN-18-nonalign}
			\centering
			\includegraphics[width=0.21\linewidth]{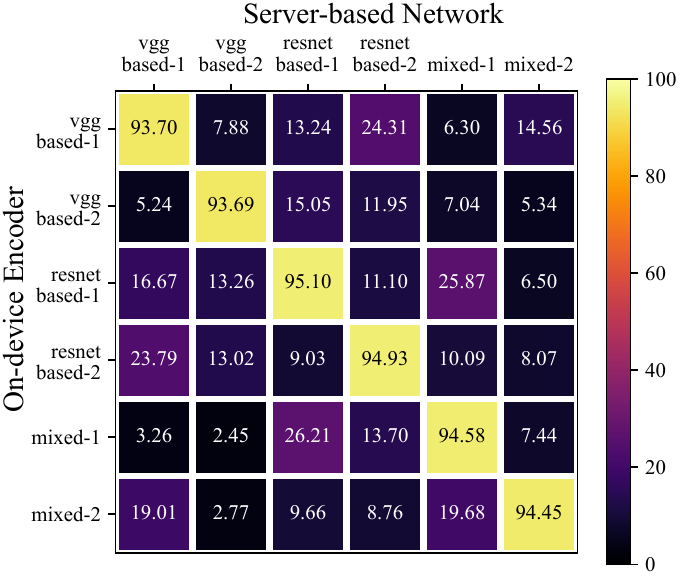}
		}
        \subfloat[FT ($\textrm{SNR}=18\textrm{dB}$)]{
        \label{subfig: SVHN-18-FT}
			\centering
			\includegraphics[width=0.18\linewidth]{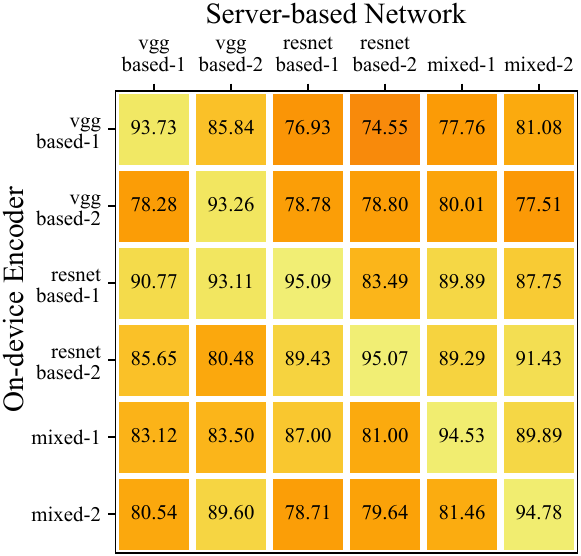}
		}
		\subfloat[OS-LS ($\textrm{SNR}=18\textrm{dB}$)]{
			\centering
			\includegraphics[width=0.18\linewidth]{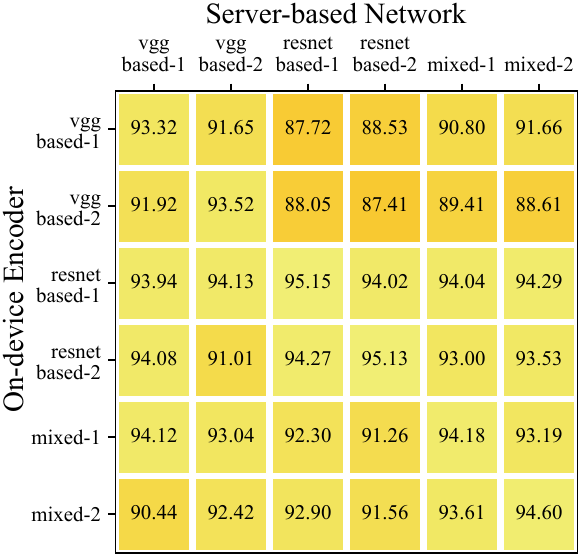}
		}
		\subfloat[OS-MMSE ($\textrm{SNR}=18\textrm{dB}$)]{
			\centering
			\includegraphics[width=0.18\linewidth]{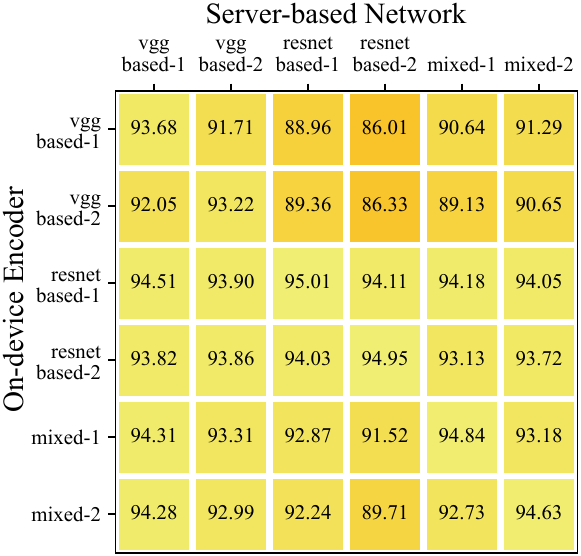}
		}
        \subfloat[OS-MLP ($\textrm{SNR}=18\textrm{dB}$)]{
			\centering
			\includegraphics[width=0.18\linewidth]{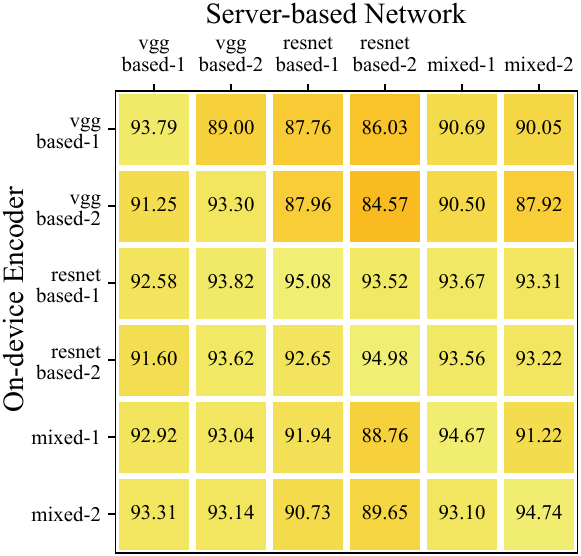}
		}
		\caption{Performance of cross-model task-oriented communications on SVHN dataset for the evaluated methods: (a) Non-Align, (b) FT, (c) OS-LS, (d) OS-MMSE, and (e) OS-MLP under AWGN channels with $\text{SNR}= 6\textrm{dB}$, and the evaluated methods: (f) Non-Align, (g) FT, (h) OS-LS, (i) OS-MMSE, and (j) OS-MLP under AWGN channels with $\text{SNR}= 6\textrm{dB}$.
		}
		\label{fig:SVHN}
	\end{figure*}
We first justify Assumption~\ref{ass-1} and Assumption~\ref{ass-2} by visualizing the feature vectors of SVHN datasets encoded by the on-device encoders with distinct architectures, each trained independently. As illustrated in Fig.\,\ref{fig:ab-align}, feature vectors are colored based on their labels. In particular, the red, green, and blue points signify data points from three randomly selected classes. Notably, we observe that the colored data points maintain fixed relative positions with slight variations in rotation and scaling, which demonstrates invariance in feature spaces up to linear transformations.

We further visualize the relative representations generated by on-device feature alignment methods across different neural network architectures. As illustrated in Fig.\,\ref{fig:ab-rel}, the relative representations show a higher degree of similarity than the original encoded features, facilitating alignment between them. This observation demonstrates the relative representations in on-device alignment can be readily recognized and directly leveraged by independent server-based neural networks.
\subsubsection{Cross-Model Inference with Distinct Trainings}
\begin{table}
	\centering
	\caption{cross-model performance for the task-oriented communication systems trained from distinct training processes with evaluated methods. }
	\label{tab: independent_training}
				\begin{tabular}{llcc|cc}
					\toprule
     \multirow{2}{*}{\textbf{Architecture}} & \multirow{2}{*}{\textbf{Approach}} & \multicolumn{2}{c|}{\textbf{SVHN}} & \multicolumn{2}{c}{\textbf{CIFAR-10}}\\
                    && WI & Hyperp & WI & Hyperp\\
					\midrule
					\multirow{7}{*}{\textbf{VGG-based}}  &Original   & {93.61} &  {93.52} &  {90.18}    & {90.08}  \\ 
				 & \cellcolor[gray]{0.8}{Non-Align}     & \cellcolor[gray]{0.8}{5.23} &  \cellcolor[gray]{0.8}{5.67} & \cellcolor[gray]{0.8}{10.52} &  \cellcolor[gray]{0.8}{11.05} \\ 
                    \cmidrule{2-6}
                    &FT & 79.94& 84.24 & 81.21 & 76.95\\
                    &OS-LS    &\underline{92.08}& {92.19}& \underline{87.34} & \underline{87.66} \\
                    &OS-MMSE & \textbf{92.36} & \textbf{92.52} & \textbf{87.90} & \textbf{88.13}\\
                    &OS-MLP & 90.96 & 91.28 & 86.86 &86.31 \\
                    &On-Device& 92.06 & \underline{92.45} & 85.94 & 85.96\\
                    \midrule
					\multirow{7}{*}{\textbf{ResNet-based}}  &Original   &95.17 & 95.25& 92.02 &91.91    \\ 
				&\cellcolor[gray]{0.8}{Non-Align} & \cellcolor[gray]{0.8}{8.90}  &\cellcolor[gray]{0.8}{9.15} & \cellcolor[gray]{0.8}{22.85}         & \cellcolor[gray]{0.8}{19.06}     \\ 
                    \cmidrule{2-6}
                    &FT &83.98 & 90.59 & 85.72 & 87.45\\
                    &OS-LS    &\textbf{94.35}& {94.10}& 87.90& {89.21} \\
                    &OS-MMSE &{94.21} & \textbf{94.48} & \textbf{90.85} & 86.81\\
                    &OS-MLP & 93.68 & 94.05 &\underline{90.18} &\textbf{90.06} \\
                    &On-Device &\underline{94.24} &  \underline{94.13} & 89.83 & \underline{89.95}\\
                    \midrule
					\multirow{7}{*}{\textbf{Mixed}}  &Original    &94.64 & 94.77& 91.29  &91.52  \\ 
				  & \cellcolor[gray]{0.8}{Non-Align}    & \cellcolor[gray]{0.8}{19.34} & \cellcolor[gray]{0.8}{19.66} &\cellcolor[gray]{0.8}{5.32}&\cellcolor[gray]{0.8}{3.58} \\ 
                    \cmidrule{2-6}
                    &FT & 89.67 & 78.63 & 83.16 & 80.88\\
                    &OS-LS  & 93.29&\underline{93.36}& \textbf{89.14}& 86.88   \\
                    &OS-MMSE  & \textbf{93.45} &\textbf{93.55} &88.92 & \textbf{89.17}\\
                    &OS-MLP & \underline{93.41} & 93.03&  \underline{89.02} &\underline{88.12} \\
                    &On-Device& 92.33 & 92.47 & 80.42 & 81.86\\
					\bottomrule
				\end{tabular}	

\end{table}
We first evaluate the performance of the proposed feature alignment methods in the cross-model task-oriented communication scenarios where each system is optimized from independent training processes. The independent training processes involve random factors of weight initialization and hyperparameters of learning rate $lr \in [4\times 10^{-4}, 1\times10^{-3}]$, denoted as \emph{WI} and \emph{Hyperp}, respectively. Table~\ref{tab: independent_training} presents the cross-model inference accuracy of the evaluated methods for independent-trained task-oriented communication systems with different neural network architectures. 

As shown in Table.\,\ref{tab: independent_training}, the accuracy of direct cross-model inference without any alignment is lower than 20\%. With server-based feature alignment, the inference accuracy has a significant increase. Furthermore, all the proposed server-based and on-device alignment methods outperform the fine-tuning method with a significant increase of over 10\%. 
These results demonstrate the effectiveness of the proposed feature alignment techniques.
\subsubsection{Cross-Model Inference with Different Architectures}
\begin{figure*}
		\centering
		\subfloat[Non-Align ($\textrm{SNR}=6\textrm{dB}$)]{
  \label{subfig: CIFAR-6-nonalign}
			\centering
			\includegraphics[width=0.21\linewidth]{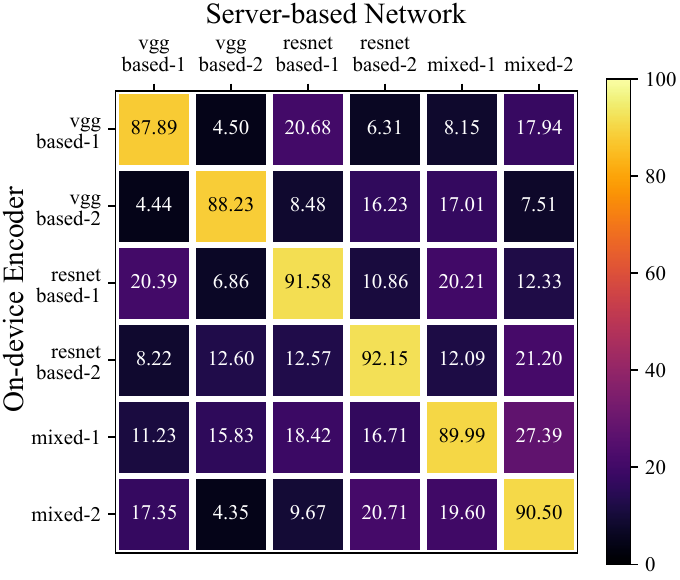}
		}
        \subfloat[FT ($\textrm{SNR}=6\textrm{dB}$)]{
			\centering
			\includegraphics[width=0.18\linewidth]{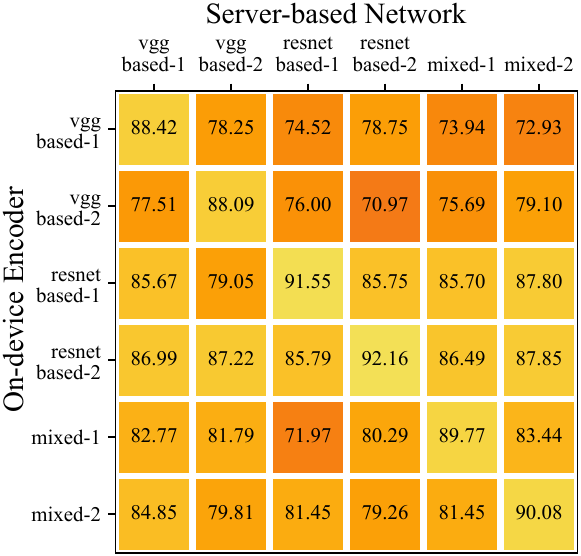}
		}
		\subfloat[OS-LS ($\textrm{SNR}=6\textrm{dB}$)]{
			\centering
			\includegraphics[width=0.18\linewidth]{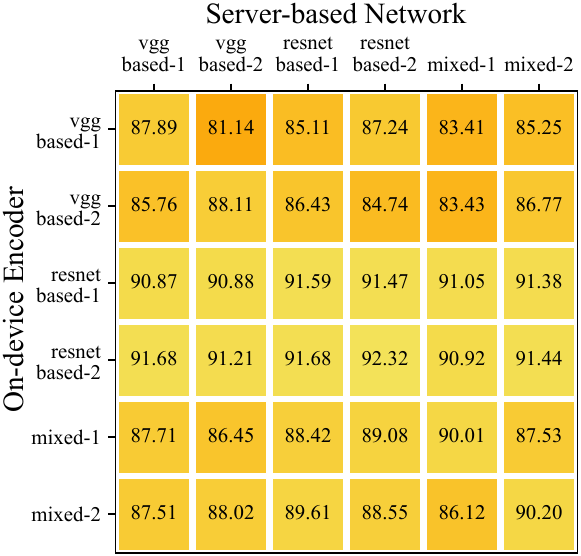}
		}
		\subfloat[OS-MMSE ($\textrm{SNR}=6\textrm{dB}$)]{
			\centering
			\includegraphics[width=0.18\linewidth]{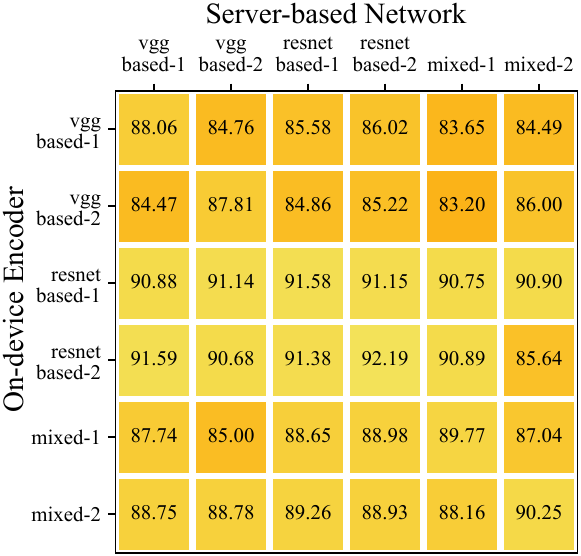}
		}
        \subfloat[OS-MLP ($\textrm{SNR}=6\textrm{dB}$)]{
			\centering
			\includegraphics[width=0.18\linewidth]{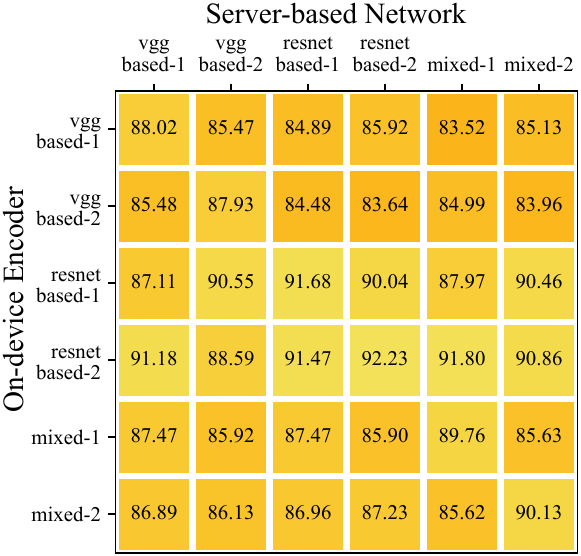}
		}

        \subfloat[Non-Align ($\textrm{SNR}=18\textrm{dB}$)]{
        \label{subfig: CIFAR-18-nonalign} 
			\centering
			\includegraphics[width=0.21\linewidth]{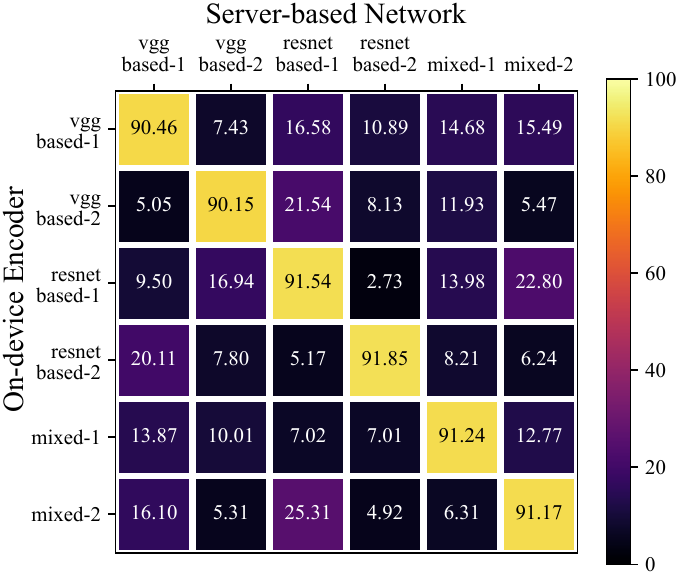}
		}
        \subfloat[FT ($\textrm{SNR}=18\textrm{dB}$)]{
			\centering
			\includegraphics[width=0.18\linewidth]{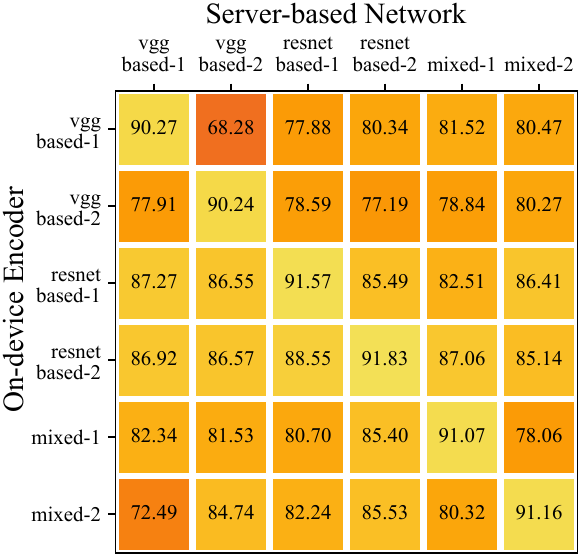}
		}
		\subfloat[OS-LS ($\textrm{SNR}=18\textrm{dB}$)]{
			\centering
			\includegraphics[width=0.18\linewidth]{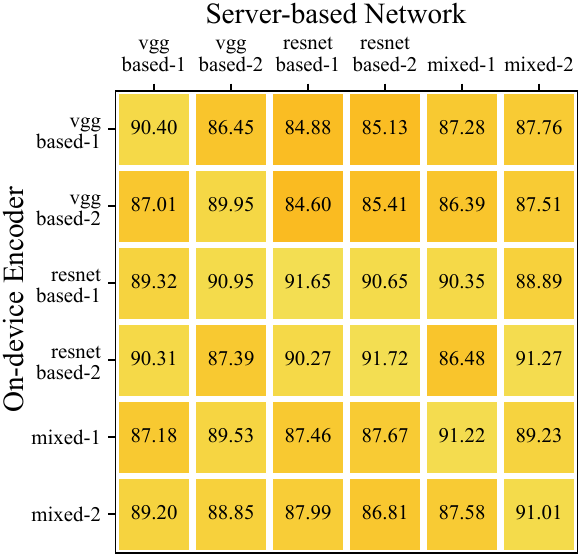}
		}
		\subfloat[OS-MMSE ($\textrm{SNR}=18\textrm{dB}$)]{
			\centering
			\includegraphics[width=0.18\linewidth]{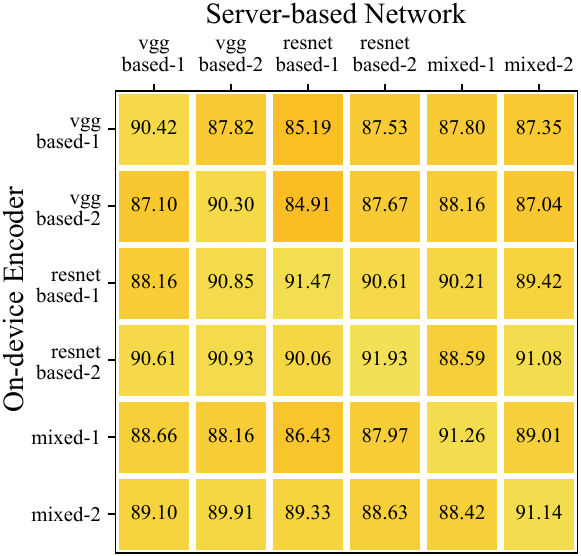}
		}
        \subfloat[OS-MLP ($\textrm{SNR}=18\textrm{dB}$)]{
			\centering
			\includegraphics[width=0.18\linewidth]{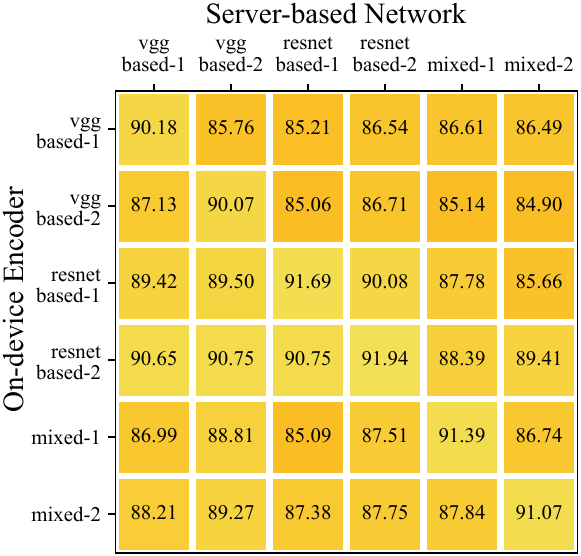}
		}
		\caption{Performance of cross-model task-oriented communications on CIFAR-10 dataset for the evaluated methods: (a) Non-Align, (b) FT, (c) OS-LS, (d) OS-MMSE, and (e) OS-MLP under AWGN channels with $\text{SNR}= 6\textrm{dB}$, and the evaluated methods: (f) Non-Align, (g) FT, (h) OS-LS, (i) OS-MMSE, and (j) OS-MLP under AWGN channels with $\text{SNR}= 6\textrm{dB}$.
		}
		\label{fig:CIFAR}
	\end{figure*}
We further evaluate the performance of feature alignment in cross-architecture scenarios where the on-device encoders and server-based networks have different neural network architectures. For each architecture, we independently train two pairs of on-device encoders and server-based networks. For instance, we train two VGG-based architectures, named \emph{vgg-based-1} and \emph{vgg-based-2}.
The numerical results for SVHN and CIFAR-10 datasets are illustrated in Fig.\,\ref{fig:SVHN} and Fig.\,\ref{fig:CIFAR} with the combinations of trained on-device encoders and server-based networks, respectively.

\begin{figure}
		\centering
		\subfloat[$\textrm{SNR}=6\textrm{dB}$]{
			\centering
			\includegraphics[width=0.43\linewidth]{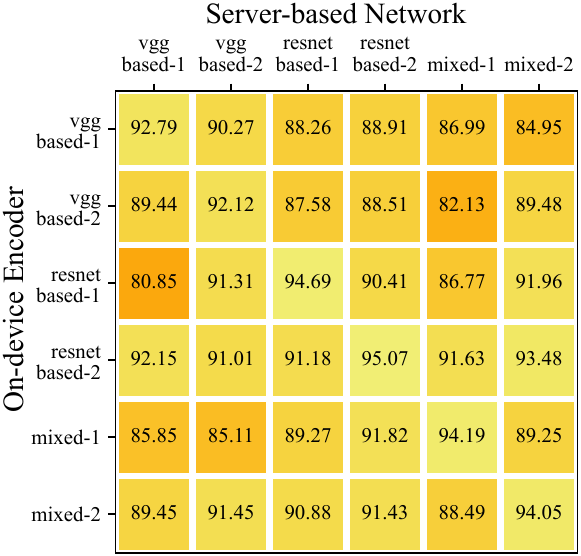}
		}
        \subfloat[$\textrm{SNR}=18\textrm{dB}$]{
			\centering
			\includegraphics[width=0.43\linewidth]{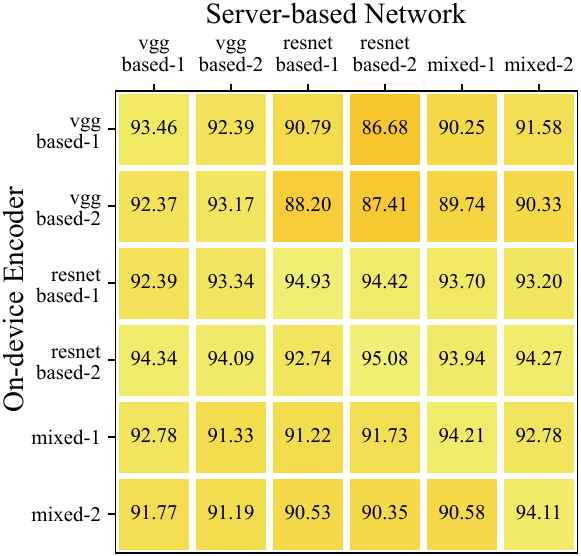}
		}

        \subfloat[$\textrm{SNR}=6\textrm{dB}$]{
			\centering
			\includegraphics[width=0.43\linewidth]{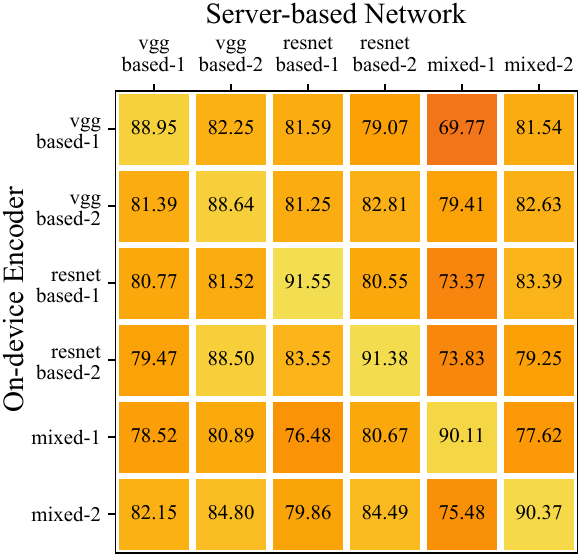}
		}
        \subfloat[$\textrm{SNR}=18\textrm{dB}$]{
			\centering
			\includegraphics[width=0.43\linewidth]{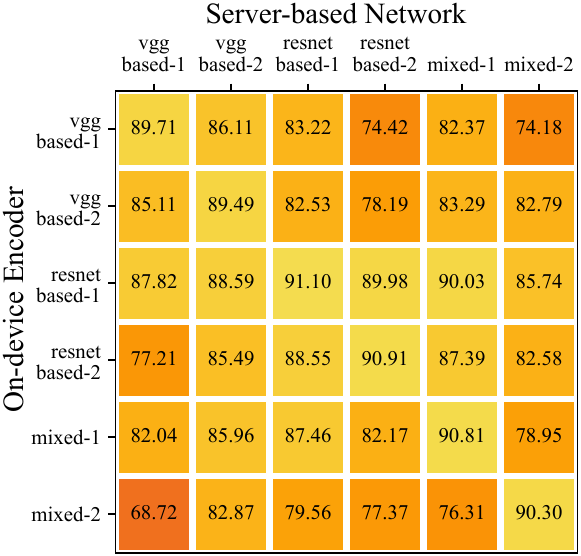}
		}
		\caption{Performance of the on-device feature alignment on SVHN dataset over AWGN channels with (a) $\text{SNR}= 6\textrm{dB}$ and (b) $\text{SNR}= 18\textrm{dB}$, and on CIFAR-10 dataset over AWGN channels with (c) $\text{SNR}= 6\textrm{dB}$ and (d) $\text{SNR}= 18\textrm{dB}$
		}
		\label{fig: zero-shot}
	\end{figure}

The results of cross-architecture inference without alignment are presented in Fig.\,\ref{subfig: SVHN-6-nonalign} and Fig.\,\ref{subfig: SVHN-18-nonalign}. The diagonal entries of these figures exhibit satisfactory accuracy, while other entries commonly fall below 20\%. These observations demonstrate the serious performance loss due to cross-architecture inference. 
As illustrated in Fig.\,\ref{subfig: SVHN-6-FT} and Fig.\,\ref{subfig: SVHN-18-FT}, fine-tuning can improve the performance of cross-architecture inference, there are still noticeable performance gaps compared to the diagonal entries. Furthermore, the proposed server-based feature alignment methods achieve the best performance and have small performance gaps compared to the optimal performance. A similar conclusion can be obtained in Fig.\,\ref{fig:CIFAR}. As for on-device alignment, the experimental findings depicted in Fig.~\ref{fig: zero-shot} illustrate the performance is close to that of the server-based alignment. These simulation results demonstrate the effectiveness of the proposed alignment methods across diverse neural network architectures.
\subsubsection{Feature Alignment under Different Channel SNRs}
\begin{figure}
		\centering
        \subfloat[SVHN]{
        \label{subfig: lines-one-shot-SVHN}
			\centering
			\includegraphics[width=0.485\linewidth]{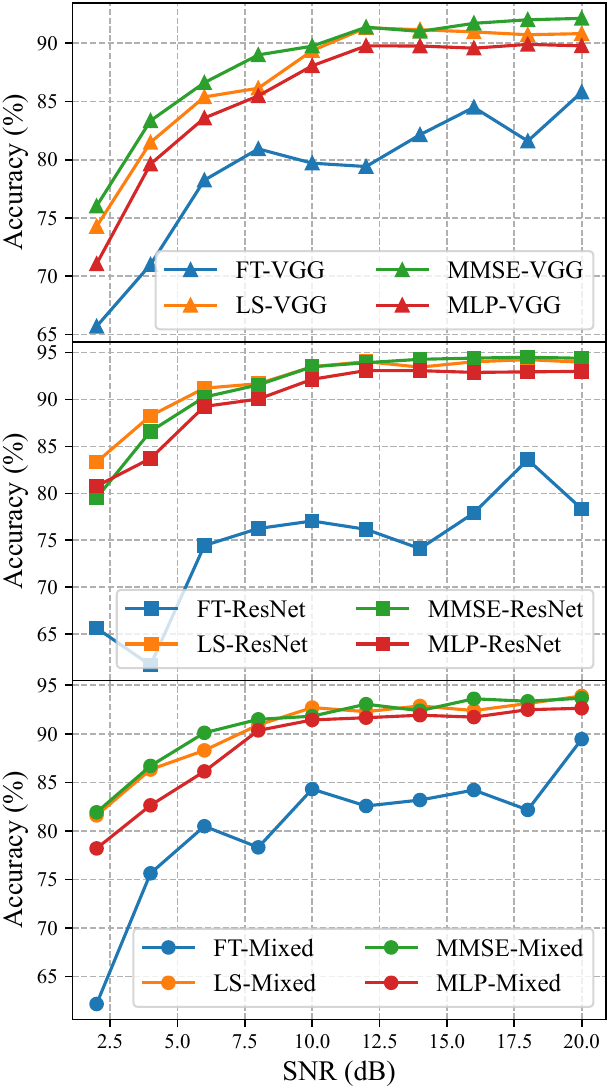}
		}
        \hspace{-10pt}
        \subfloat[CIFAR-10]{
        \label{subfig: lines-one-shot-CIFAR}
			\centering
			\includegraphics[width=0.485\linewidth]{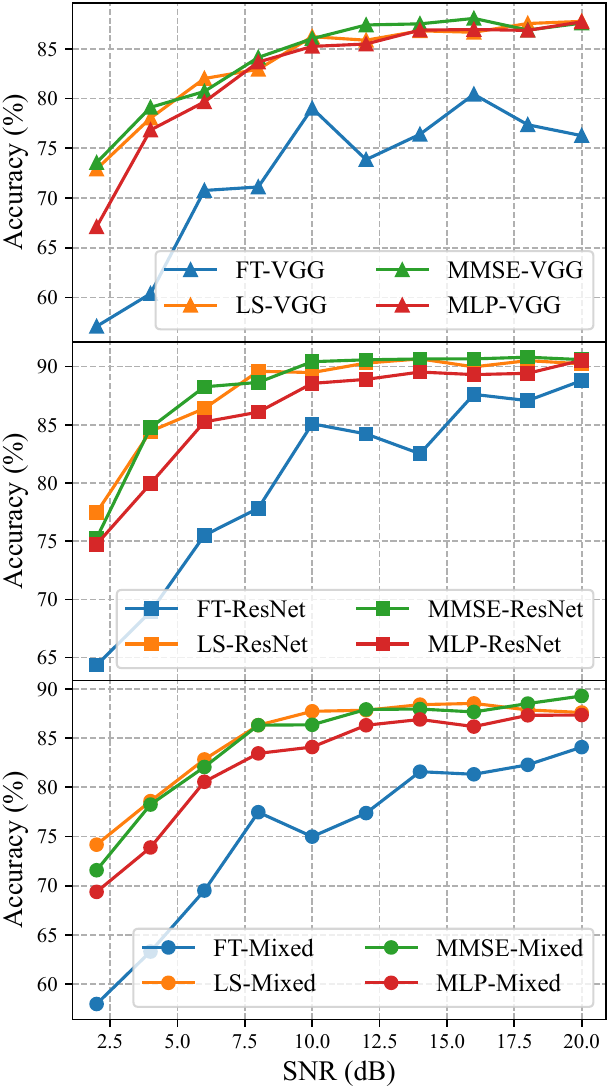}
		}
		\caption{The cross-model task-oriented communication performance for evaluated server-based feature alignment methods with various neural network architectures under the AWGN channels with $\textrm{SNR} \in [2 \textrm{dB}, 20\textrm{dB}]$ on (a) SVHN and (b) CIFAR-10 datasets.
		}
  \label{fig: lines-one-shot}
\end{figure}
\begin{figure}
		\centering
        \subfloat[SVHN]{
        \label{subfig: lines-relative-SVHN}
			\centering
			\includegraphics[width=0.75\linewidth]{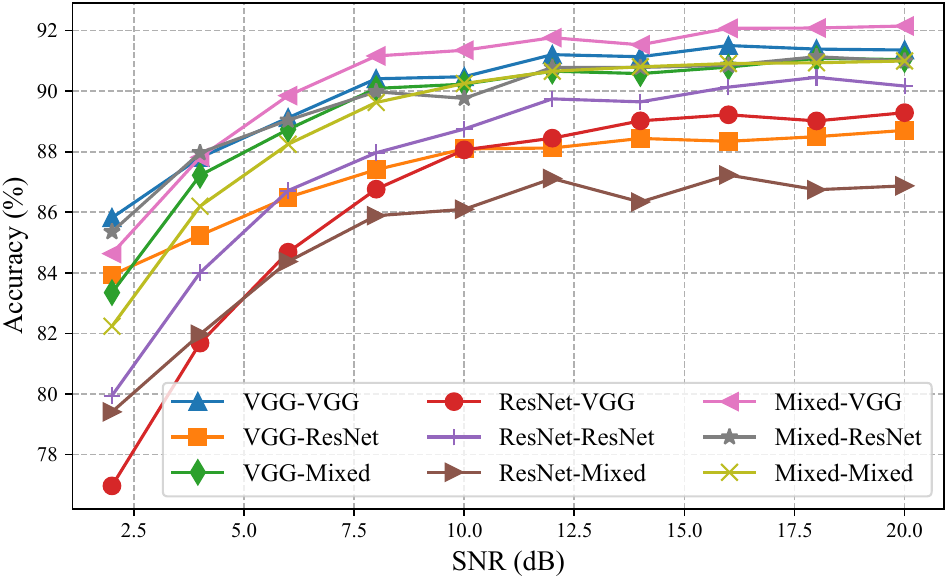}
		}
  
        \subfloat[CIFAR-10]{
        \label{subfig: lines-relative-CIFAR}
			\centering
			\includegraphics[width=0.75\linewidth]{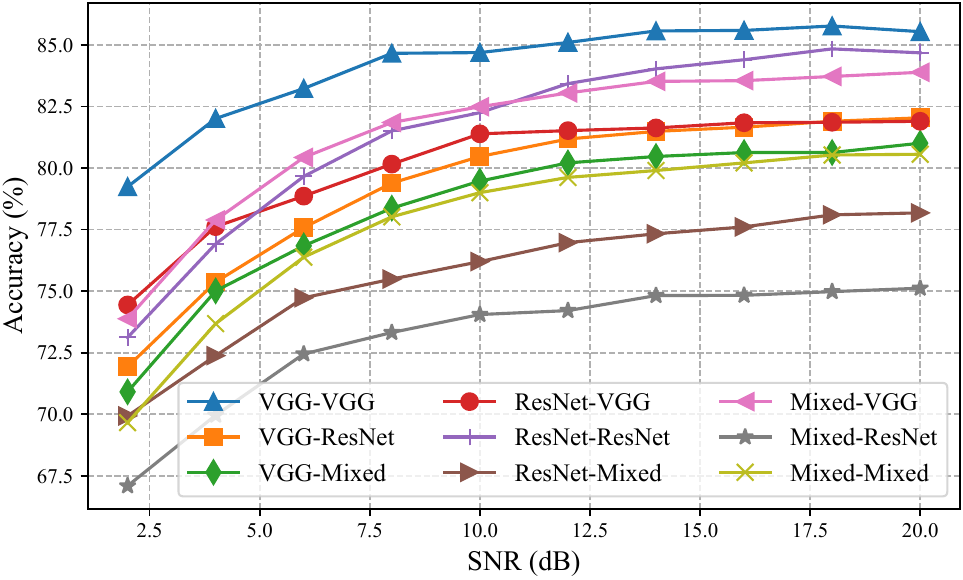}
		}
		\caption{Performance of the proposed on-device feature alignment under the AWGN channels with $\textrm{SNR} \in [2 \textrm{dB}, 20\textrm{dB}]$ for the cross-model task-oriented communication with different neural network architectures.
		}
  \label{fig: lines-zero-shot}
\end{figure}

We further conduct the experiments for the proposed methods with various SNRs. The experimental results for server-based alignment methods on SVHN and CIFAR-10 datasets are shown in Fig.~\ref{subfig: lines-one-shot-SVHN} and Fig.~\ref{subfig: lines-one-shot-CIFAR}, respectively. It can be observed that the fine-tuning exhibits unstable inference performance over the changing channel SNRs, and all the proposed server-based alignment methods exhibit consistent performance over the changing SNRs. The performance of the proposed on-device alignment with changing SNRs is presented in Fig.~\ref{fig: lines-zero-shot}. Although the cross-model task-oriented communication with different server-based encoders and on-device networks have different inference performances, we observe the consistent performance with different SNRs under all the scenarios. Note that the above edge inference systems are trained and tested under the same channel SNR. There is still a performance gap if there is a channel SNR mismatch between training and test phrases. To bridge this gap, recent studies have explored the integration of additional architectures like attention modules and Hypernetworks~\cite{xu2021wireless,10621322}, which can be utilized.

\subsection{Runtime Analysis}
In the above experiments, we have demonstrated the effectiveness of the proposed server-based and on-device feature alignment methods for cross-model task-oriented communication. In this subsection, we analyze the latency, runtime, and memory cost for feature alignment methods. The latency of the edge inference for each image is determined by the dimension of the transmitted feature vector $d$. In the simulation, we set up $d=16$ and thus have a low latency of $1.67$ ms at a symbol rate of 9,600 Baud. As the dimension of the relative representation is equal to the number of anchor data $n_{\tau}$, the transmission latency of relative representations is $3.33$ ms because $n_{\tau}=32$ in our experiments.
 \begin{table}[t]
		\caption{Runtime of the evaluated methods with different architectures.}
		\label{tab: runtime}
  \centering
			\begin{tabular}{l|ccc}
				\toprule
				  Architecture & VGG-based & ResNet-based & Mixed \\
				\midrule
     FT & 970.52 \ms& 521.38 \ms & 583.82 \ms \\
     OS-MLP& 25.99 \ms & 25.64 \ms & 25.16 \ms\\
OS-LS & 0.36 \ms & 0.39 \ms & 0.37\ms \\
OS-MMSE&  0.60 \ms & 0.56 \ms & 0.57 \ms \\
on-device & 0 \ms & 0 \ms & 0 \ms \\
				\bottomrule
		\end{tabular}
	\end{table}
\begin{table}[t]
		\caption{Additional Memory and Computation Overhead of the evaluated alignment methods.}
		\label{tab: memory}
  \centering
			\begin{tabular}{l|ccc}
				\toprule
				 Method & \#Params & Size & FLOPs\\
				\midrule
    FT & $10,432$ & $40.75$ KB & $20,480$ \\
OS-LS & $256$ & $1$ KB & $512$\\
OS-MMSE& $256$ & $1$ KB & $512$\\
OS-MLP& $256$ & $1$ KB & $512$ \\
on-device & $0$ & $0$ KB & $96n_{\tau}$ \\
				\bottomrule
		\end{tabular} 
	\end{table}

The runtime of different alignment methods is presented in Table~\ref{tab: runtime}. The proposed methods have significantly lower latency than fine-tuning. Note that the on-device alignment method is $0$ ms. That is because the proposed on-device method transmits the relative representations that can be directly recognized by other edge servers. However, the on-device alignment has additional transmission latency if the number of anchor data is larger than the dimension of encoded feature vectors. Furthermore, the additional memory and computational overhead of different feature alignments are shown in Table~\ref{tab: memory}. Compared to fine-tuning, the size and computation overhead of the server-based alignment methods is significantly small. Note that the on-device feature alignment does not introduce additional memory requirements but introduces additional computation overhead at the edge devices that depend on the number of anchor data $n_{\tau}$. In summary, the proposed server-based and on-device alignment methods exhibit memory efficiency on both edge devices and edge servers.

%% file: secs/6_ending.tex
\section{Conclusion}\label{sec: conc}
{In this work, we investigated cross-model task-oriented communication in edge inference systems for real-time CV applications. To facilitate cross-model inference, we formulated the feature alignment problem within system models that incorporate independent task-oriented models and shared anchor data. By leveraging the linear invariance and angle-preserving properties of visual semantic features, we developed both server-based and on-device feature alignment approaches for cross-model task-oriented communication. Extensive experiments demonstrated the effectiveness of our methods and validated their capability for real-time applications.}

The proposed feature alignment approaches offer promising solutions for the model compatibility issue in cross-model task-oriented communication. {However, to deploy edge AI systems in practical environments, several key research areas must be addressed. First, systems need to be adaptable to varying real-time conditions. Second, ensuring compatibility across diverse models is crucial for seamless integration. Protecting user privacy is essential, requiring robust safeguards for sensitive data. Additionally, enhancing resilience against adversarial attacks is vital for maintaining system integrity. Finally, improving transparency in AI decision-making is necessary to build user trust. Tackling these challenges is critical for the successful implementation of secure and efficient edge AI systems.}

%% file: secs/7_appendix.tex
\section{Proofs of Proposition 1}
\label{appendix a}
Before proving the difference of upper bounds, we adopt the linear transformation $\M$ between feature spaces and establish the equality $\bE_{p_{\bphi_1}(\z_1|\x)}[\log p_{\btheta_2}(\y|\M\tz_1]=\bE_{p_{\bphi_2}(\z_2|\x)}[p_{\btheta_2}(\y|\z_2+\M\bepsilon)]$ by changing the variable from $\z_1$ to $\z_2$. It can be proven by,
\begin{align}
    &\bE_{p_{\bphi_1}(\z_1|\x)}[\log p_{\btheta_2}(\y|\M\tz_1]\\
    =&\bE_{p_{\bphi_1}(\z_1|\x)}[\log p_{\btheta_2}(\y|\M(\z_1+ \bepsilon)]\\
    =&\bE_{p_{\bphi_1}(\z_1|\x)}[\log p_{\btheta_2}(\y|\z_2+ \M\bepsilon)]\\
    =&\int p_{\bphi_1}(\z_1|\x)\log p_{\btheta_2}(\y|\z_2+\M\bepsilon) d\z_1\\
    =&\int p_{\bphi_2}(\z_2|\x)|\M | \log p_{\btheta_2}(\y|\z_2+\M\bepsilon) |\M|^{-1} d\z_2\\
    =&\int p_{\bphi_2}(\z_2|\x)\log p_{\btheta_2}(\y|\z_2+\M\bepsilon) d\z_2\\
    =& \bE_{p_{\bphi_2}(\z_2|\x)}[p_{\btheta_2}(\y|\z_2+\M\bepsilon)].\label{eq:app_1}
\end{align}
Then, the upper bound of the absolute difference between $\mathcal{L}_{\text{OS}} ( \M, \bphi_1, \btheta_2)$ and $\mathcal{L}(\bphi_2, \btheta_2)$ can be written as,
\begin{align}
    &\left| \mathcal{L}_{\text{OS}} ( \M, \bphi_1, \btheta_2) - \mathcal{L}(\bphi_2, \btheta_2)\right|\\
    =& |\bE_{p(\x, \y)p(\bepsilon)p_{\bphi_1}(\z_1|\x)}[\log p_{\btheta_2}(\y|\M\tz_1)]-\notag\\ 
    & \quad \quad \quad \quad \quad \quad \ \ \mathbb{E}_{p(\x,\y)p(\bepsilon)p_{\bphi_2}(\z_2|\x)} [\log p_{\btheta_2}(\y|\tz_2)]|\\
    \overset{(a)}{=}& |\bE_{p(\x, \y)p(\bepsilon)p_{\bphi_2}(\z_2|\x)}[\log p_{\btheta_2}(\y|\z_2 + \M\bepsilon)]-\notag\\
    & \quad \quad \quad \quad \quad \mathbb{E}_{p(\x,\y)p(\bepsilon)p_{\bphi_2}(\z_2|\x)} [\log p_{\btheta_2}(\y|\z_2 + \bepsilon)]|\\
    \overset{(b)}{\leq}& \bE_{p(\x, \y)p(\bepsilon)p_{\bphi_2}(\z_2|\x)}[|\log p_{\btheta_2}(\y|\z_2 + \M\bepsilon)-\notag\\
    & \quad \quad \quad \quad \quad \quad \quad \quad \quad \quad \quad \quad \quad\log p_{\btheta_2}(\y|\z_2 + \bepsilon)|]\\
    =& \bE_{p(\x)p(\bepsilon)p_{\bphi_2}(\z_2|\x)}[\sum\limits_{\y \in \mathcal{Y}}p(\y|\x)|\log p_{\btheta_2}(\y|\z_2 + \M\bepsilon)-\notag\\
    & \quad \quad \quad \quad \quad \quad \quad \quad \quad \quad \quad \quad \quad\log p_{\btheta_2}(\y|\z_2 + \bepsilon)|]\\
    \overset{(c)}{\leq} & \bE_{p(\x)p(\bepsilon)p_{\bphi_2}(\z_2|\x)}[\| f_{\btheta_2}(\z_2 + \M\bepsilon) - f_{\btheta_2}(\z_2 + \bepsilon)\|_{\infty}\\
    \overset{(d)}{\leq} & \bE_{p(\x)p(\bepsilon)p_{\bphi_2}(\z_2|\x)}[\rho\| (\z_2 + \M\bepsilon) - (\z_2 + \bepsilon)\|_{\infty}\\
    = & \bE_{p(\x)p(\bepsilon)p_{\bphi_2}(\z_2|\x)}[\rho\| ( \M-\mathbf{I})\bepsilon\|_{\infty}\\
    \overset{(e)}{\leq} & \|\M-\mathbf{I} \|_{\infty} \bE_{p(\bepsilon)}[\| \bepsilon\|_{\infty}]\\
    \overset{(f)}{\leq} &  \sigma \sqrt{2\log d} \| \M -\mathbf{I} \|_{\infty}
\end{align}
where (a) follows the equality~\eqref{eq:app_1}, (b) follows the inequality for the expectation of absolute difference, (c) comes from the fact that log probabilities $\log p_{\theta}(\mathbf{y}|\mathbf{z})$ are modeled by the output of the neural network function $f_{\theta_2}(\mathbf{z})$ and the expected value is less than the maximum output value, (d) follows the $\rho$-Lipschitz continuity for neural network function $f_{\btheta_2}(\cdot)$, (e) follows the Cauchy–Schwarz inequality, and (f) follows the next conclusion provided.

For $d$ i.i.d. Gaussian random variables $\epsilon_j \sim N(0, \sigma^2), 1\leq j \leq d$, it is easy to obtain that the expectation of maximum absolute value is upper bounded by,
\begin{align}
    \bE [\max\limits_{1\leq j \leq d}|\epsilon_i|] \leq \sigma \sqrt{2\log d}.
\end{align}

\section{Derivation of MMSE Estimator} \label{appendix b}
Let us obtain the estimator $\hMSE = \Tilde{\mathbf{Z}}_{\tau,2} \mathbf{A}^*$ that minimize the estimation MSE. Note $\bE[\Upsilon^T\Upsilon] = d\sigma^2 \mathbf{I}$ and let $\mathbf{R}_{\M}=\bE[\M^T\M]$ deontes correlation matrix of $\M$. This estimation error can be written as
\begin{align}
    \textrm{MSE} & = \mathbb{E} [\|\M -  (\M(\Tilde{\bZ}_{\tau, 1} -\Upsilon) + \Upsilon)\mathbf{A}\|^2_F]\\
    & = \bE [\Tr\{ (\M -  (\M(\Tilde{\bZ}_{\tau, 1} -\Upsilon) + \Upsilon)\mathbf{A})^T( \M -\\
    & \qquad \qquad \qquad \qquad \quad \ (\M(\Tilde{\bZ}_{\tau, 1} -\Upsilon) + \Upsilon)\mathbf{A})\}\\
    &=\Tr\{\mathbf{R}_{\M}\} - \Tr\{\mathbf{R}_{\M}\Tilde{\bZ}_{\tau, 1}\mathbf{A} \} -\Tr\{\mathbf{A}^T\Tilde{\bZ}_{\tau, 1}^T\mathbf{R}_{\M} \}+ \notag \\
    & \quad \ \Tr\{\mathbf{A}^T(\Tilde{\bZ}_{\tau, 1}^T \mathbf{R}_{\M} \Tilde{\bZ}_{\tau, 1} + n_{\tau}\sigma^2 \mathbf{R}_{\M} + n_{\tau}\sigma^2 \mathbf{I})\mathbf{A} \}.
\end{align}
As the prior distribution of each entry of $\M$ is i.i.d. Gaussian variable $N(0, 1/\sqrt{d})$, we have the correlation matrix $\mathbf{R}_{\M} = \mathbf{I}$. The optimal $\mathbf{A}$ is obtained by $\partial \varepsilon/ \partial \mathbf{A}=0$, that is
\begin{align}
    \mathbf{A}^* & = (\Tilde{\bZ}_{\tau, 1}^T \Tilde{\bZ}_{\tau, 1} + 2n_{\tau}\sigma^2 \mathbf{I})^{-1}\Tilde{\bZ}_{\tau, 1}^T.
\end{align}
Then, the MMSE estimator is given by,
\begin{align}
    \hMSE & = \Tilde{\bZ}_{\tau, 2}(\Tilde{\bZ}_{\tau, 1}^T \Tilde{\bZ}_{\tau, 1} + 2n_{\tau}\sigma^2 \mathbf{I})^{-1}\Tilde{\bZ}_{\tau, 1}^T.
\end{align}